\documentclass[11pt]{searcheyes}

\usepackage{enumitem}
\usepackage{tabularx}
\usepackage{float}
\usepackage{amsmath}

\definecolor{goldcolor}{RGB}{252,200,100}
\definecolor{silvercolor}{RGB}{240,230,200}
\definecolor{top1orange}{RGB}{244,165,66}
\definecolor{top2orange}{RGB}{247,231,206}

\newcommand{\coco}{\textbf{CoCo}}

\newfontfamily\arxivkeepbytesansmedium[Path=./]{ByteSans-Medium.ttf}
\newfontfamily\arxivkeepbytesansbold[Path=./]{ByteSans-Bold.ttf}

\title{\gradtext{CoCo}: Code-as-CoT for Text-to-Image Preview and Rare Concept Generation}

\author[1,2]{Haodong Li}
\author[2,*]{Chunmei Qing}
\author[3]{Huanyu Zhang}
\author[1]{Dongzhi Jiang}
\author[2]{\\Yihang Zou}
\author[1]{Hongbo Peng}
\author[1]{Dingming Li}
\author[1]{Yuhong Dai}
\author[2]{ZePeng Lin}
\author[4]{\\Juanxi Tian}
\author[2]{Yi Zhou}
\author[1]{Siqi Dai}
\author[1]{Jingwei Wu}
\author[1,*]{Pheng-Ann Heng}

\affiliation[1]{The Chinese University of Hong Kong}
\affiliation[2]{South China University of Technology}
\affiliation[3]{Institute of Automation, Chinese Academy of Sciences}
\affiliation[4]{Nanyang Technological University}

\contribution[*]{Corresponding author.}

\abstract{%
Recent advancements in Unified Multimodal Models (UMMs) have significantly advanced text-to-image (T2I) generation, particularly through the integration of Chain-of-Thought (CoT) reasoning. However, existing CoT-based T2I methods largely rely on abstract natural-language planning, which lacks the precision required for complex spatial layouts, structured visual elements, and dense textual content. In this work, we propose \coco{} (\textbf{Co}de-as-\textbf{Co}T), a code-driven reasoning framework that represents the reasoning process as executable code, enabling explicit and verifiable intermediate planning for image generation. Given a text prompt, \coco{} first generates executable code that specifies the structural layout of the scene, which is then executed in a sandboxed environment to render a deterministic draft image. The model subsequently refines this draft through fine-grained image editing to produce the final high-fidelity result. To support this training paradigm, we construct \textbf{CoCo-10K}, a curated dataset containing structured draft--final image pairs designed to teach both structured draft construction and corrective visual refinement. Empirical evaluations on StructT2IBench, OneIG-Bench, and LongText-Bench show that \coco{} achieves improvements of +68.83\%, +54.8\%, and +41.23\% over direct generation, while also outperforming other generation methods empowered by CoT. These results demonstrate that executable code is an effective and reliable reasoning paradigm for precise, controllable, and structured text-to-image generation.%
}

\date{March 2026}
\correspondence{\email{mickyhimself4@gmail.com}}
\checkdata[Code]{\url{https://github.com/micky-li-hd/CoCo}}

\begin{document}
\maketitle

\section{Introduction}
\label{sec:intro}

\begin{figure}[tp]
    \centering
    \includegraphics[width=\linewidth]{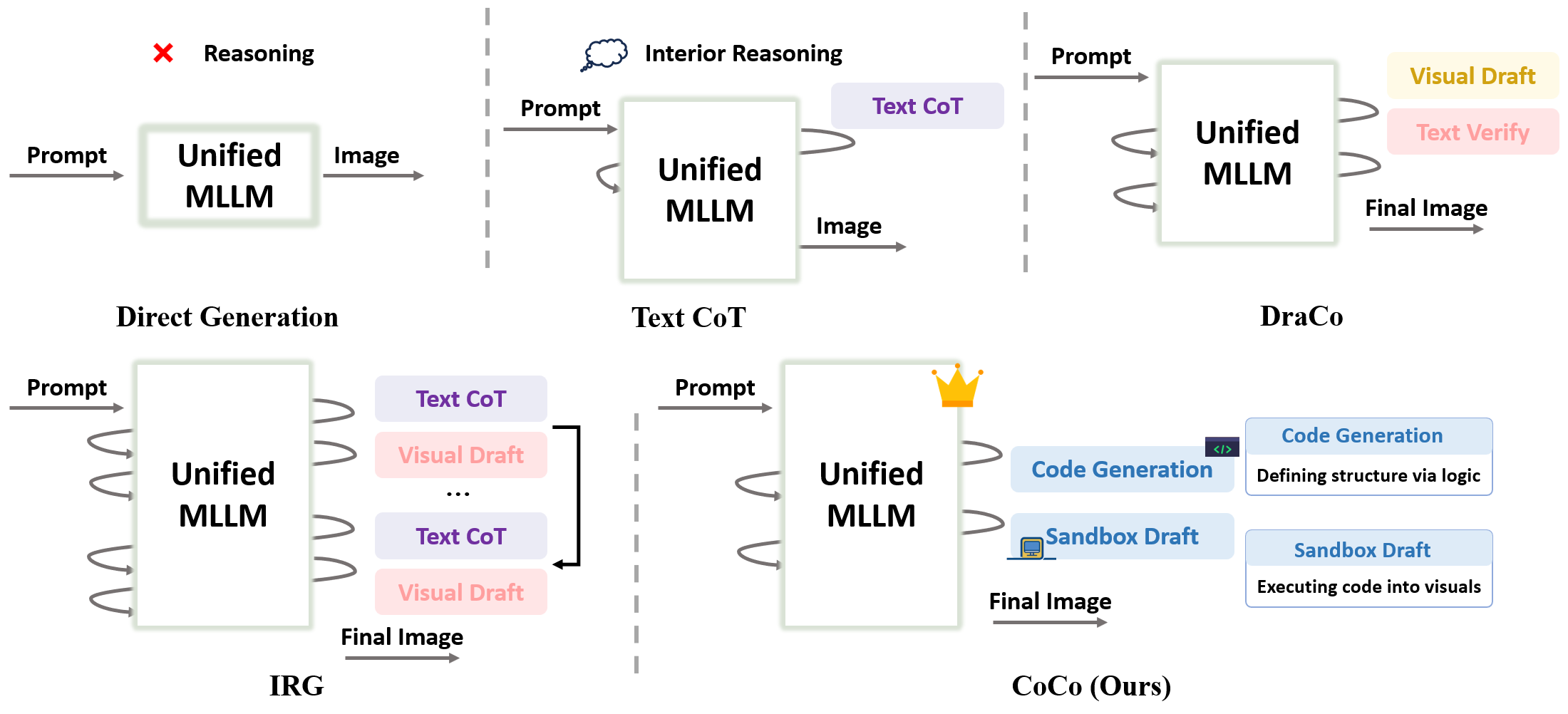}
    \caption{\textbf{Comparison of reasoning paradigms for text-to-image generation.}
    (a) Direct generation without explicit reasoning.
    (b) Text-only Chain-of-Thought (CoT) reasoning prior to image synthesis.
    (c) Textual CoT with an intermediate visual draft.
    (d) Multi-turn reasoning with iterative textual and visual drafts.
    (e) \textbf{\coco{}}: our approach generates executable code as a structured visual draft that explicitly specifies object layouts and attributes, which is then rendered and refined to produce the final image.}
    \label{fig:intro}
\end{figure}

Unified multimodal large language models (MLLMs)~\citep{deng2025emerging,wu2024janus,wu2024vila,zhou2024transfusion,chen2025blip3,xiao2025omnigen,team2024chameleon,liquid,qu2024tokenflow,lin2025uniworld} have recently emerged as a compelling paradigm that jointly models visual understanding and generation within a single architecture, as exemplified by Bagel~\citep{deng2025emerging}, EMU3~\citep{wang2024emu3}, and Janus~\citep{wu2024janus}. This consolidation gives rise to emergent capabilities beyond those of single-task models—most notably, the ability to process interleaved image-text inputs and synthesize images conditioned on complex multimodal instructions~\citep{wu2025omnigen2,xin2025lumina,zhang2026vibe-benchmark,lyu2025understanding}.

Concurrently, chain-of-thought (CoT) reasoning~\citep{wei2022chain,kojima2022large} has demonstrated significant improvements across a broad spectrum of tasks, spanning mathematical problem solving~\citep{amini2019mathqa,hendrycksmath2021,Lu2023MathVistaEM,zhang2024mathverse}, visual reasoning~\citep{yue2023mmmu,jiang2025mme,chen2025r1v,zhan2025visionr1,meng2025mm,zhang2025scaling,zhang2025latent}, and multi-agent systems~\citep{mai2025agent,wei2025webagent}. Recent efforts have sought to extend CoT to text-to-image (T2I) generation~\citep{guo2025can,deng2025emerging,zhang2025reasongen,pan2025focusdiff,gu2025improving,lyu2025realrag}, yet these approaches remain limited in their exploitation of the unified MLLM architecture. Image-Gen-CoT~\citep{guo2025can}, for instance, employs a reward model to assess generation quality at early decoding steps, effectively reducing the model to a pure image generator. A subsequent line of work~\citep{deng2025emerging} prepends textual reasoning to the generation process, demonstrating promising improvement.

However, natural-language planning is inherently too abstract to specify the precise spatial layouts, structural elements, and fine-grained visual attributes demanded by complex prompts.
This limitation becomes particularly evident in structured visual domains such as \textit{scientific diagrams} and images containing dense \textit{textual content}, where accurate layout, symbol placement, and semantic consistency are critical.
In practice, current generation models often struggle with prompts such as ``2D plot of \(y = x^2\)'', frequently producing incorrect structures or illegible text due to the lack of explicit visual grounding during reasoning.
In contrast, expressing the reasoning process in executable code enables a more precise and verifiable form of planning.
Unlike natural-language descriptions, code can explicitly encode spatial layouts, structural constraints, and textual placements in a deterministic and executable manner.
Once executed, the code produces a draft image that concretely instantiates the model's reasoning outcome, making the planned structure directly observable and verifiable.
This executable reasoning process transforms abstract planning into an explicit visual scaffold, allowing the model to inspect the rendered draft and perform targeted refinements.
As a result, the final image can be progressively improved from the draft, achieving higher visual fidelity and stronger semantic alignment with the input prompt.
This raises a natural question: \textbf{\textit{Can executable code serve as a more precise and verifiable form of CoT to produce an explicit draft image that guides structured and text-intensive image generation?}}

Building on this motivation, we introduce \textbf{\coco{}} (\textbf{Co}de‑as‑\textbf{Co}T), a unified reasoning–generation framework that leverages executable code as an explicit intermediate representation for T2I synthesis.
Rather than only relying on abstract natural-language descriptions, as demonstrated in Fig.~\ref{fig:intro}, CoCo requires the model to express its planning process as code that encodes spatial layouts, structural constraints, and object relationships in a deterministic manner.
The generated code will be executed within a sandboxed environment to produce a draft image that faithfully reflects the intended structure.
This draft then serves as a concrete visual scaffold on which the model performs fine‑grained image editing to produce the final high‑fidelity result.

To realize this code‑driven reasoning paradigm, the unified MLLM must not only generate executable code but also manipulate the draft images with precise, targeted edits. However, existing datasets provide neither the necessary supervision for code generation nor the fine‑grained correction signals required for this workflow. To address this gap, we construct \coco{}‑10K, a carefully curated dataset built around three atomic correction capabilities. For each capability, we design a rigorous data‑synthesis pipeline that fully leverages the complementary strengths of a powerful MLLM~\citep{gemini} and image‑editing models~\citep{bananapro}, ensuring both high‑quality sketch code and semantically aligned Draft Image–Final Image pairs.

Building upon Bagel~\citep{deng2025emerging}, we conduct extensive experiments to assess the effectiveness of our code‑driven paradigm. Results show that \coco{} delivers substantial improvements over Bagel~\citep{deng2025emerging}, achieving a 68.83\% gain on StructT2IBench \citep{zhuo2025factualitymattersimagegeneration} and further surpassing text‑CoT‑based approaches by 64.48\%. Moreover, \coco{} exhibits robust generalization on more demanding benchmarks such as LongText‑Bench \citep{geng2025xomnireinforcementlearningmakes} and OneIG‑Bench~\citep{chang2025oneigbenchomnidimensionalnuancedevaluation}. Representative visualizations of \coco{}'s outputs are provided in Fig.~\ref{fig:vis2}.

In summary, our contributions are as follows:
\begin{itemize}
    \item \textbf{Code-as-CoT reasoning for structured T2I generation.} We propose \coco{}, a novel reasoning–generation framework that leverages executable code as Chain-of-Thought.

    \item \textbf{A Code-as-CoT training dataset.} We introduce CoCo-10K, a curated dataset containing Text–Code pairs and Text–Draft–Final image triplets, enabling models to jointly learn executable layout planning and draft-guided image refinement.

    \item \textbf{Extensive empirical validation on structured and text-intensive T2I tasks.}
    Extensive experiments on StructT2IBench, LongText-Bench, and OneIG-Bench demonstrate that CoCo substantially outperforms existing methods, especially on tasks requiring precise layouts, complex text rendering, and structured visual generation.
\end{itemize}

\section{Related Work}

\subsection{Unified Multimodal Models}

UMMs aim to unify cross-modal understanding and generation, yet strong understanding often fails to yield equally strong native generation. Existing designs fall into two paradigms: pure autoregressive models that jointly predict text and visual tokens over interleaved sequences~\citep{chen2025janus, cui2025emu3, tong2025metamorph} and hybrid models that combine autoregressive language modeling with diffusion-based image synthesis, either within a unified backbone~\citep{xie2024show, zhao2024monoformer} or via modular routing and sparse experts~\citep{shi2024lmfusion, liang2024mixture, deng2025emerging}, with related guidance schemes such as Diffusion Forcing~\citep{chen2024diffusion}.
Beyond architecture, personalization~\citep{nguyen2025yo, an2025unictokens, zhong2026unified} becomes an interesting application for UMMs.
Furthermore, self-improvement methods convert self-generated signals into training objectives~\citep{yu2025guided, zhou2024calibrated, wang2025unified,lyu2024unibind}; for UMMs, SRUM derives internal rewards from understanding~\citep{jin2025srum}, and UniRL jointly optimizes understanding and generation~\citep{mao2025unirlselfimprovingunifiedmultimodal}.
Recent advances further explore efficiency and self-supervision: DeepGen 1.0~\citep{wang2026deepgen} leverages Stacked Channel Bridging (SCB) and a three-stage training strategy to enable lightweight models (5B) to outperform much larger counterparts on reasoning benchmarks, while UniCorn~\citep{han2026unicorn} introduces a ``proposer-solver-critic'' self-play framework that converts internal understanding into explicit generation signals without external data.
However, constrained by generation-side training data and the lack of targeted training strategies, UMMs tend to underperform on structured visual synthesis and complex text rendering.

\subsection{Multimodal Reasoning}

Recent breakthroughs in multimodal large language models lie in the continuously evolving reasoning paradigms~\citep{luo2025ursa, peng2025lmm, shen2025vlm, tong2025delving, wu2025visualquality, zhang2025perl,li20251+,lideepscan,li2026videococo,tian2026auto,yan2026proact,wei2026perceptionrubrics} and more unified foundation models~\citep{chen2025blip3, guo2025can, xie2025show, yang2025hermesflow}.
Starting from simple image understanding, the long-chain-of-thought (LongCoT) paradigm trained with reinforcement learning substantially strengthened multimodal reasoning~\citep{guo2025deepseek,guo2025seed1,team2025kimi}.
OpenAI-o3~\citep{team2025openai} brings visual outcomes into LongCoT, pioneering a new ``thinking with images'' paradigm of interleaved multimodal reasoning.
Meanwhile, architectures are converging toward unified designs that integrate text and image inputs/outputs~\citep{chen2025janus, xie2025reconstruction}, making interleaved reasoning within a single framework a compelling direction. Mogao~\citep{liao2025mogao} highlights the potential of interleaved generation under unified architectures, while T2I-R1~\citep{jiang2025t2i} and Bagel~\citep{deng2025emerging} explore ``thinking before generating'' for interleaved text--image synthesis. In parallel, efficient region-level understanding is achieved by PAM~\citep{lin2025perceive}, which unifies segmentation, recognition, and captioning via a semantic perceiver and parallel decoding.
However, existing approaches primarily optimize textual instructions during reasoning for image generation.
Recently, Huang et al.~\citep{huang2025interleaving} propose the Interleaving Reasoning Generation (IRG) framework, which adopts a multi-round ``text-to-image-to-text-to-image'' workflow to iteratively refine outputs through visual feedback.
In contrast, CoCo introduces executable code as an explicit and structured intermediate representation, enabling controllable draft construction prior to image synthesis rather than post-hoc refinement.
This fundamental difference allows CoCo to directly encode spatial layouts and structural constraints, providing deterministic and interpretable guidance for complex generation tasks.

\subsection{Benchmarks for Text-to-Image Generation}

Evaluating text-to-image (T2I) models requires benchmarks that probe diverse capabilities. Early benchmarks primarily focus on prompt–image alignment~\citep{yu2022scaling, hu2023tifa, huang2023t2i, ghosh2023geneval, hu2024ella, li2024genai, wu2024conceptmix, huang2025t2i, wei2025tiif}. For example, GenEval~\citep{ghosh2023geneval} verifies object co-occurrence, position, count, and color through object detection, while TIIF-Bench~\citep{wei2025tiif} extends this paradigm to more complex instruction following with a fine-grained VLM-based evaluation protocol covering text rendering, long instructions, and designer-level constraints. More recent benchmarks further examine structured and text-intensive visual generation. StructT2IBench~\citep{zhuo2025factualitymattersimagegeneration} evaluates factual accuracy in structured image synthesis involving charts, mathematical figures, and tables using a multi-round QA-based metric. OneIG-Bench~\citep{chang2025oneigbenchomnidimensionalnuancedevaluation} expands evaluation to multilingual text rendering, stylized generation, and compositional scenarios, while LongText-Bench~\citep{geng2025xomnireinforcementlearningmakes} specifically targets the accurate rendering of extended textual content within images. Recent efforts also explore more complex capabilities, such as interactive and editing-based evaluation, including GEBench~\citep{li2026gebench}, which assesses multi-step interaction in GUI environments, VIBE~\citep{zhang2026vibe-benchmark}, which evaluates visual instruction-driven editing, and GENIUS~\citep{an2026geniusgenerativefluidintelligence}, which examines generative fluid intelligence. Collectively, these benchmarks highlight that current generative models still face notable challenges when handling structured layouts, dense textual content, and reasoning-intensive visual generation.

This gap motivates the development of CoCo, which introduces executable code as a structured reasoning medium to bridge semantic intent and visual realization.

\section{Method}

We first introduce the structure of our adopted unified MLLM, Bagel~\citep{deng2025emerging}, in Section~\ref{method: preliminary}. Then we elaborate on the design idea and pipeline of \coco{} in Section~\ref{method: Draft-as-CoT}. Finally, we introduce the construction of \coco{}-10K in Section~\ref{method: training_dataset}.

\begin{figure*}[!t]
    \centering
    \includegraphics[width=\linewidth]{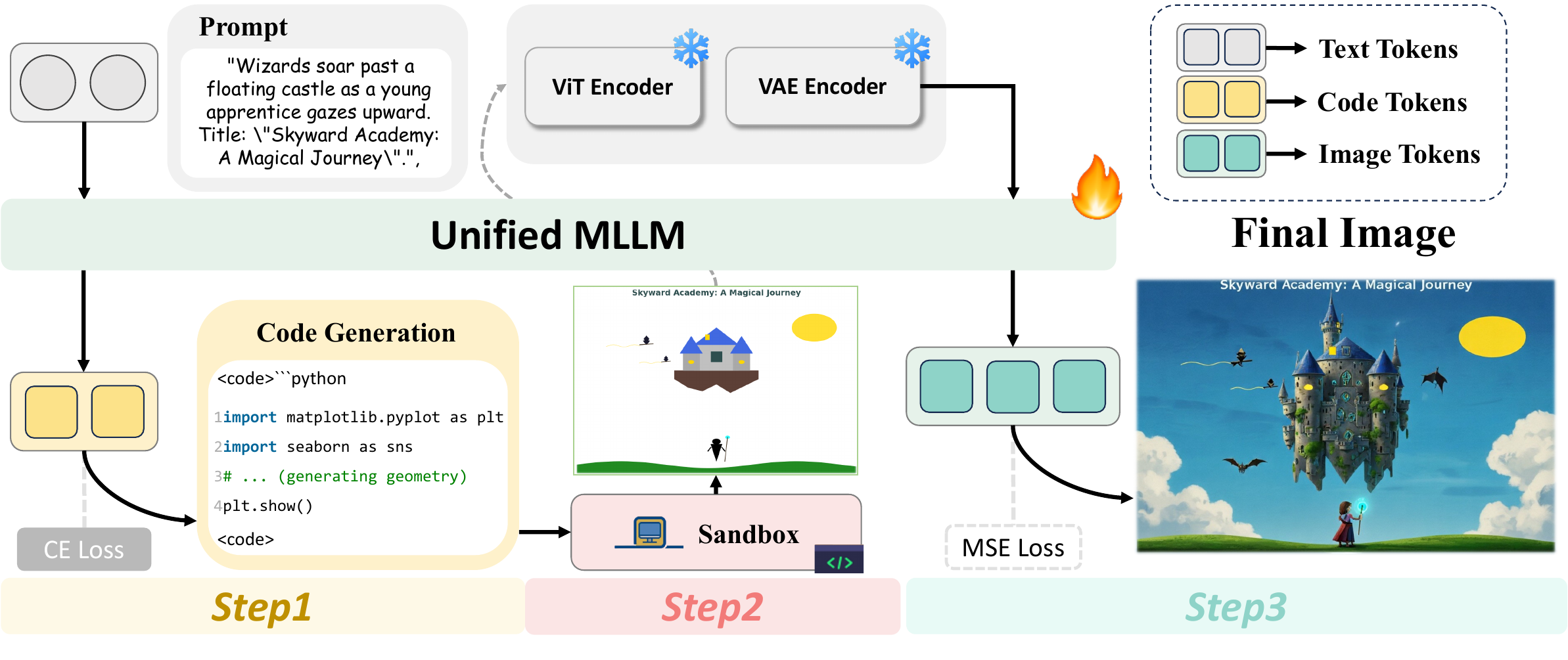}
    \caption{\textbf{Framework of \coco{}.}
    \coco{} operates in three stages: (1) generating executable code from the input prompt to explicitly specify structural layouts, (2) executing the code in a sandboxed environment to render a deterministic draft image, and (3) performing draft-guided refinement to produce the final high-quality output.
    }
    \label{fig:main}
\end{figure*}

\subsection{Preliminary}
\label{method: preliminary}
Our work builds upon Bagel~\citep{deng2025emerging}, a unified MLLM that supports both visual generation and visual understanding within a single architecture. Bagel~\citep{deng2025emerging} comprises three core components: a SigLIP ViT encoder~\citep{tschannen2025siglip} for visual understanding, a VAE encoder~\citep{kingma2013auto} for generation, and a Mixture-of-Transformer-Experts (MoT) consisting of two specialized transformer branches~\citep{vaswani2017attention}, one dedicated to VAE tokens for visual generation and the other to ViT and text tokens for understanding.
For understanding tasks, an input image is encoded by the ViT into a sequence of visual tokens, which are then processed by the transformer to autoregressively predict text tokens.
For generation tasks, Bagel~\citep{deng2025emerging} adopts Rectified Flow~\citep{lipman2022flow,liu2022flow,esser2024scaling} to decode VAE tokens into high-fidelity images.

\subsection{Code-as-CoT Framework}
\label{method: Draft-as-CoT}
The overall pipeline of \coco{} is illustrated in Fig.~\ref{fig:main}.
Given a textual prompt, our framework decomposes the generation process into three stages: \textit{code generation}, \textit{draft image rendering}, and \textit{draft-guided refinement}.
Instead of directly synthesizing the final image, \coco{} first produces executable code that explicitly encodes structural semantics, renders a draft image through code execution, and finally refines the draft to obtain the final result.

\paragraph{\textbf{Code Generation.}}
Leveraging the language reasoning capabilities of unified MLLMs,
the first stage generates executable code that serves as an explicit form of Chain-of-Thought reasoning.
Given a user prompt $p$, the model produces code $c$ that deterministically specifies the core semantic structure of the target image.
This code explicitly describes key visual elements such as spatial layouts, object relationships, textual rendering, and canvas configurations.
Instead of synthesizing a fully detailed image in a single step,
the generated code focuses on specifying the essential semantic structure of the scene.
Fine-grained visual details, such as stylistic appearance and rendering quality, are deferred to later refinement stages.

\paragraph{\textbf{Draft Image Rendering.}}
The generated code $c$ is executed in a sandbox environment to produce a draft image $I_d$.
This step instantiates the programmatic reasoning into a concrete visual representation.
The sandbox ensures safe and stable execution through restricted operations and isolated runtime environments.
The resulting draft $I_d$ captures the core semantic layout specified by the code, including object placement, textual content, and structural relationships.

\begin{figure*}[!t]
    \centering
    \includegraphics[width=\linewidth]{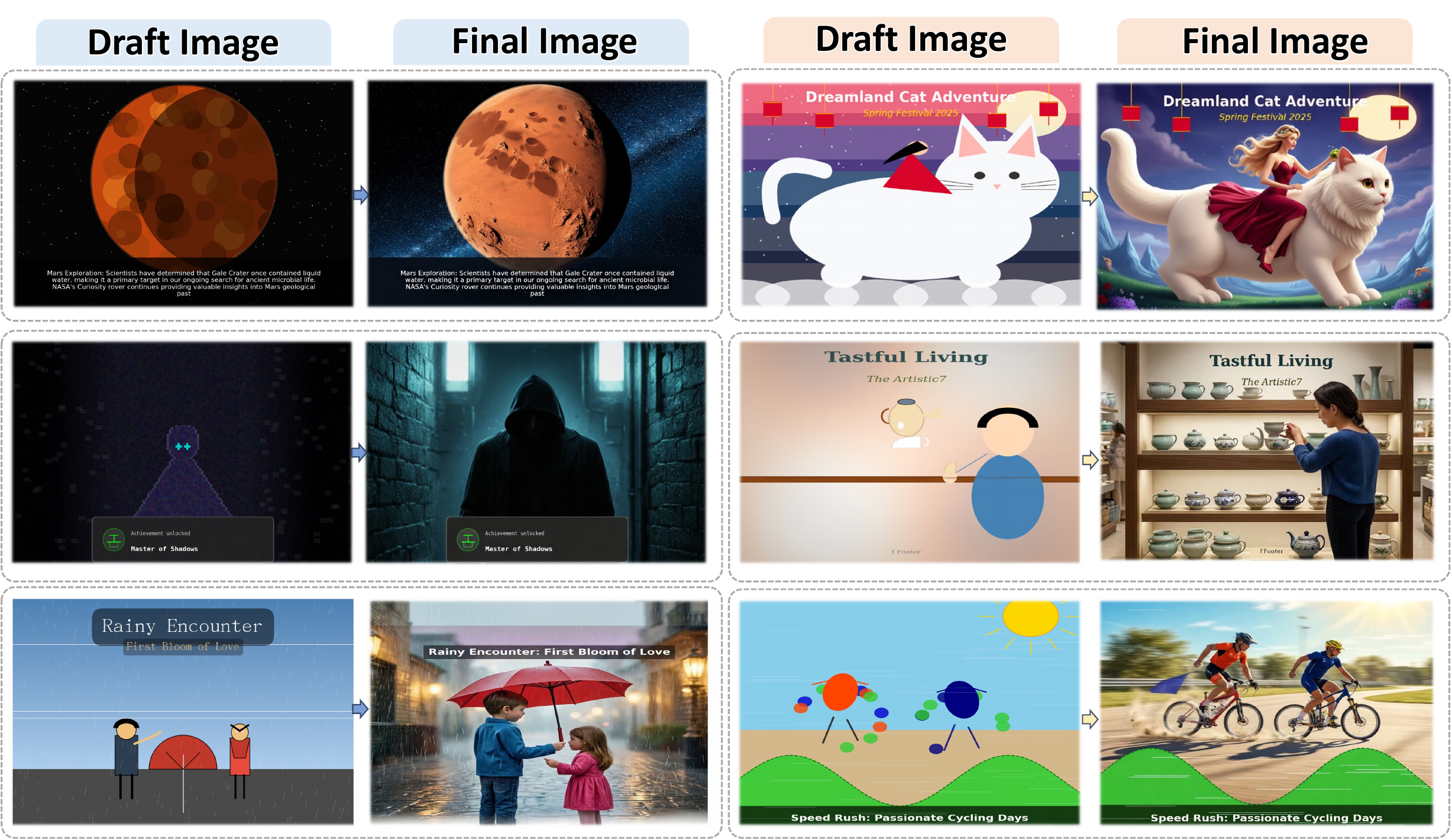}
    \caption{\textbf{Visual comparison between code-generated drafts and final results.}
    Left: draft image synthesized from code generated by \coco{}.
    Right: final image refined by \coco{} based on the draft image and input prompt.
    }
    \label{fig:vis2}
\end{figure*}

\paragraph{\textbf{Draft-Guided Refinement.}}
While the draft image provides an accurate structural scaffold, its visual appearance may remain simplistic due to programmatic rendering.
To obtain the final output, we introduce a draft-guided refinement stage that improves visual fidelity while preserving strong semantic alignment with the draft.
As illustrated in Fig.~\ref{fig:vis2}, the refined images retain the structural layout specified by the draft while significantly enhancing visual realism.
Specifically, in Bagel~\citep{deng2025emerging}, the draft image $I_d$ is encoded using both a ViT encoder and a VAE encoder before being fed back into the unified MLLM.
The ViT encoder extracts high-level semantic features for global reasoning, while the VAE encoder preserves low-level visual details for precise image editing.
Together, these representations enable the model to enhance visual realism while maintaining the structural semantics defined by the draft.

\subsection{{\bf \coco{}}-10K}
\label{method: training_dataset}

We combine UMM-based code generation with image editing capabilities to facilitate a sketch-to-image paradigm, enabling the synthesis of complex scenes through structured layout generation and subsequent refinement.
We first explain the necessity of constructing the dataset and describe the two stages of the generation pipeline for each capability. Finally, we outline how the collected data are organized for training.

\subsubsection{Necessity of Dataset Construction}
\label{dataset: necessity}
\begin{figure*}[!htb]
    \centering
    \includegraphics[width=\linewidth]{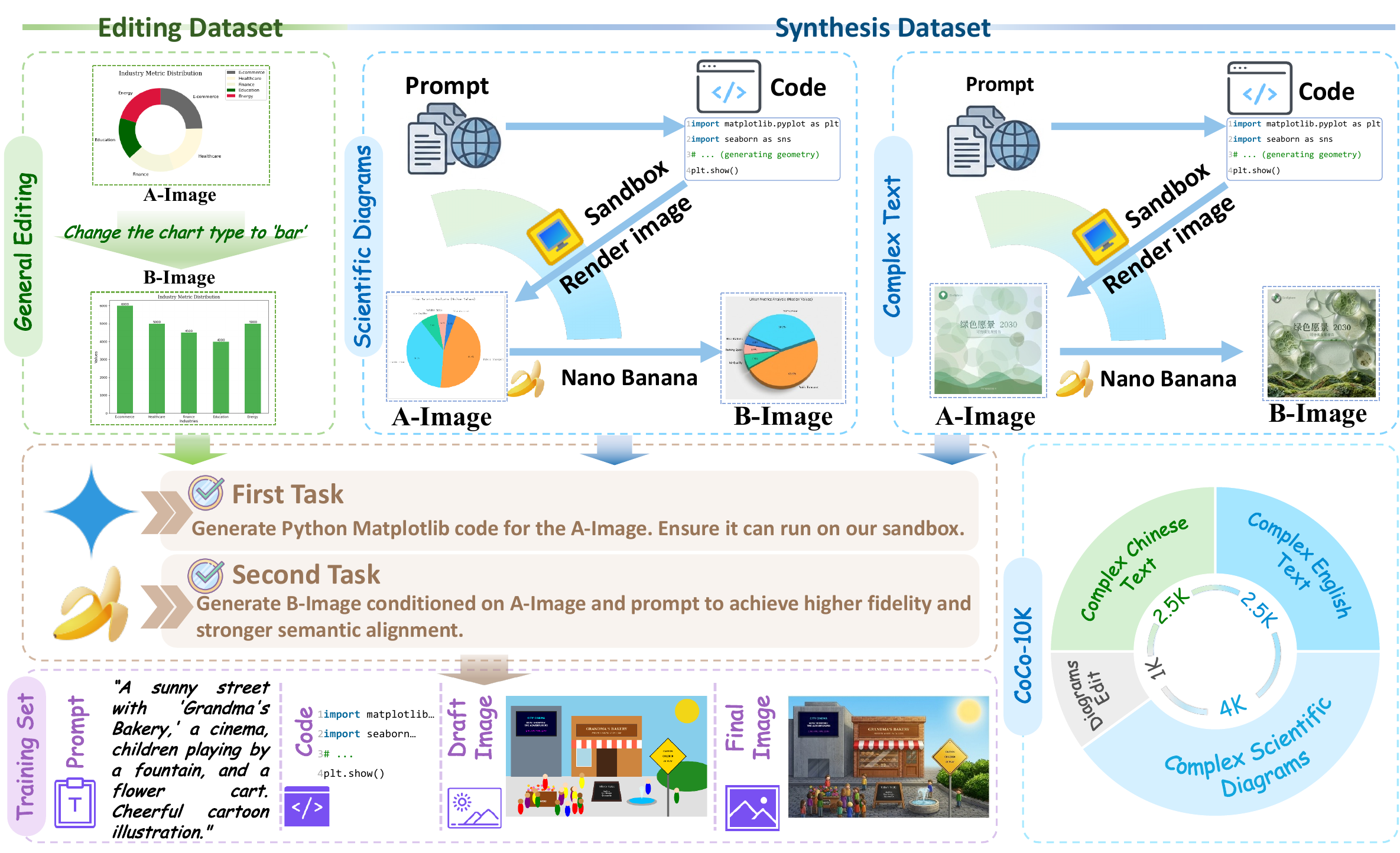}
    \caption{\textbf{Construction Pipeline and Examples of \coco{}-10K.}
        We design specialized data pipelines targeting three atomic correction capabilities: general editing, scientific diagrams, and complex text. Using Gemini-3-Pro~\citep{gemini} and Nano Banana~\citep{bananapro}, we generate code and final images from the collected prompts. The resulting data are organized into two training categories: Text-Code Pairs and Text-Draft Image–Final Image Pairs.
        }
    \label{fig:dataset}
\end{figure*}

Our zero-shot pilot study reveals two key limitations of Bagel~\citep{deng2025emerging}.
First, although the model is capable of producing code, the generated code is often non-executable and leads to low-quality outputs that lack the required precision and fidelity.
Second, the model tends to ignore the draft semantics and instead produces entirely new images, defeating the purpose of using drafts to ease the difficulty of one-shot generation.

However, existing datasets are insufficient to address these issues.
Therefore, we construct \coco{}-10K, a training dataset containing over 10K structured supervision instances.
We identify three essential atomic capabilities required for correction and design dedicated data pipelines for each of them.
Throughout the data construction process, we ensure the generation of high-fidelity sketch code while maintaining strict semantic consistency between draft images and final outputs.
This design enables the model to not only produce executable and robust code but also learn selective editing—correcting erroneous elements while preserving correct structures.

\subsubsection{Construction Details}
\label{dataset: details}

As illustrated in Fig.~\ref{fig:dataset}, our dataset consists of two complementary components: the \textit{Editing Dataset} and the \textit{Synthesis Dataset}. The Editing Dataset focuses on preserving the model's perception of structured visual elements, while the Synthesis Dataset is specifically designed for structured visual generation. In particular, the synthesis data explicitly provides paired intermediate renderings and refined outputs, which align with the two-stage generation paradigm of \coco{} and enable the model to learn both draft construction and draft-guided refinement.

\textbf{Editing Dataset.}
The Editing Dataset is constructed from StructVisuals~\citep{zhuo2025factualitymattersimagegeneration}, which contains a large collection of structured chart images. For each sample, we denote the original image as \textit{A-Image} and the corrected result as \textit{B-Image}. These pairs typically involve modifications to data values, labels, or formatting while preserving the overall chart structure. Such characteristics allow the model to maintain a strong perception of structured visual layouts and learn to perform precise corrections without disrupting the underlying structure.

\textbf{Synthesis Dataset.}
While the Editing Dataset helps preserve structural perception, generating complex structured visuals remains challenging for existing generative models, particularly for \textit{scientific diagrams} and images containing dense \textit{complex text}. To address this limitation, we construct a dedicated synthesis dataset using an automated generation pipeline that explicitly produces paired intermediate and final images.

First, we synthesize a diverse set of prompts covering scientific concepts and text-intensive visual formats such as charts, posters, infographics, and annotated diagrams. Gemini-3-Pro~\citep{gemini} is then employed to generate corresponding code that explicitly specifies the visual structure and layout. These scripts are executed within a sandbox environment to render initial programmatic visualizations, which we denote as \textit{A-Image}. As these renderings typically exhibit simple visual styles, we further apply Nano Banana~\citep{bananapro} to refine them into visually enhanced outputs conditioned on both the original prompt and \textit{A-Image}, producing the final \textit{B-Image}.

\begin{figure*}[tp]
    \centering
    \includegraphics[width=\linewidth]{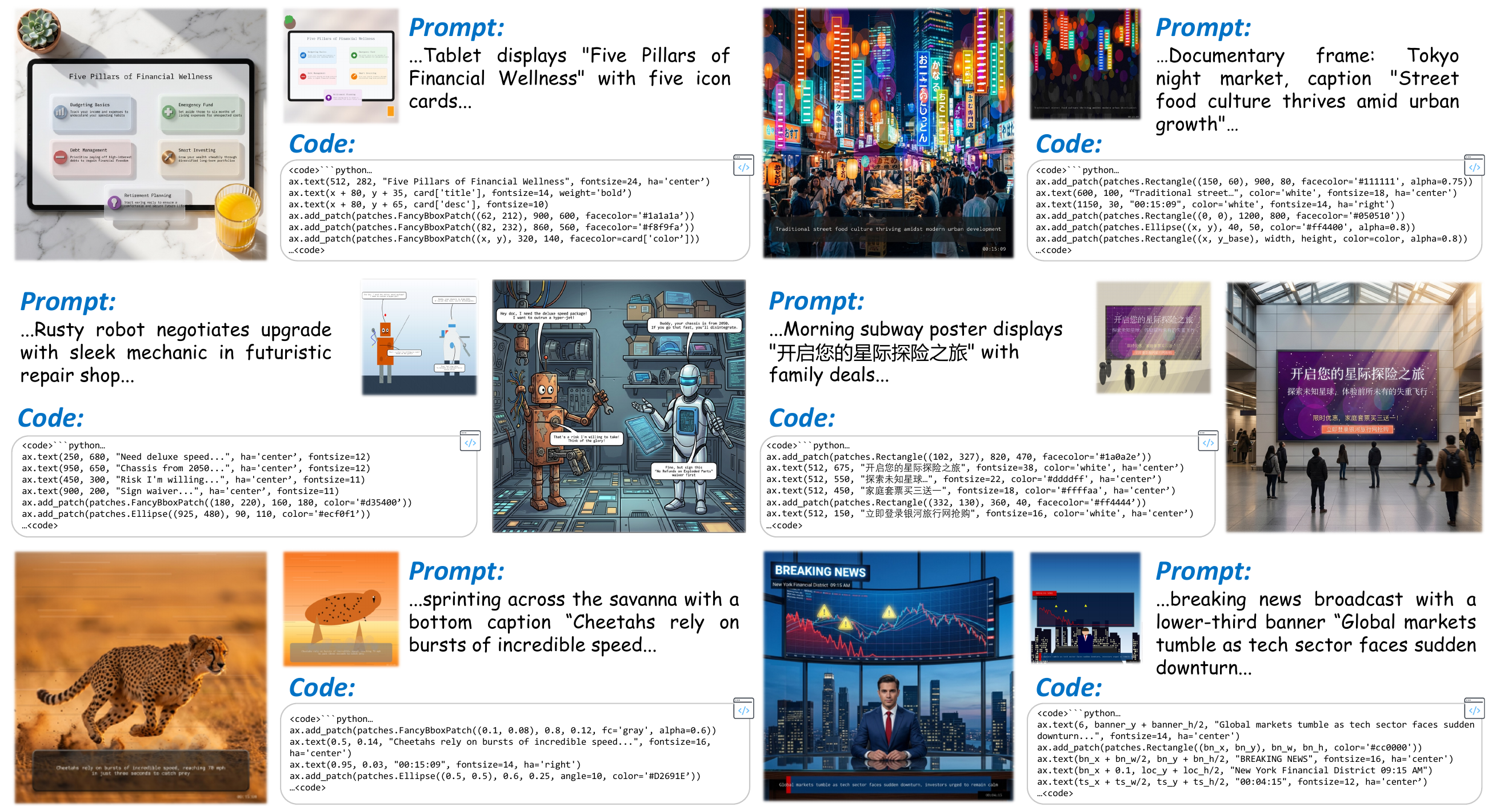}
    \caption{\textbf{Detailed Visualization of \coco{}-10K.} We present the prompt, generated code, draft image, and final image. \coco{}-10K demonstrates high fidelity and strong semantic alignment between draft and final images.}
    \label{fig:vis11}
\end{figure*}

This process naturally constructs paired supervision between executable code, its rendered draft (\textit{A-Image}), and the refined result (\textit{B-Image}), which closely mirrors the two-stage generation paradigm of \coco{}. Such alignment enables the model to learn both structured draft construction and subsequent visual refinement for complex generation tasks.
Examples from \coco{}-10K are illustrated in Fig.~\ref{fig:vis11}, showing the prompt, generated code, rendered draft image, and the corresponding refined result.

\subsubsection{Training Set}
\label{dataset: training}
Finally, we organize the prompt, code, and corresponding images into two training formats:
(1) Text–Code pairs and (2) Text–Draft Image–Final Image triplets.

\subsection{Training Loss}
\label{method: loss}
We conduct supervised fine-tuning from Bagel~\citep{deng2025emerging}. For our curated training dataset, we sequentially input the prompt tokens, the ViT feature of draft images, the verifications, and finally the noisy VAE tokens of the final image.
We only calculate token-level cross entropy loss on the code and Mean Squared Error (MSE) on the VAE tokens:
\begin{align}
\mathcal{L}_{\text{code}} &= -\frac{1}{|v|} \sum_{i=1}^{|v|}  \log(v_i), \\
\mathcal{L}_{\text{final image}} &= \mathbb{E}_{t, {x}_0, {x}_1} \left[ \left\| m(t, {x}_t) - ({x}_1 - {x}_0) \right\|^2 \right].
\end{align}

\section{Experiment}

\subsection{Experimental Setting}

\begin{figure*}[tp]
    \centering
    \includegraphics[width=\linewidth]{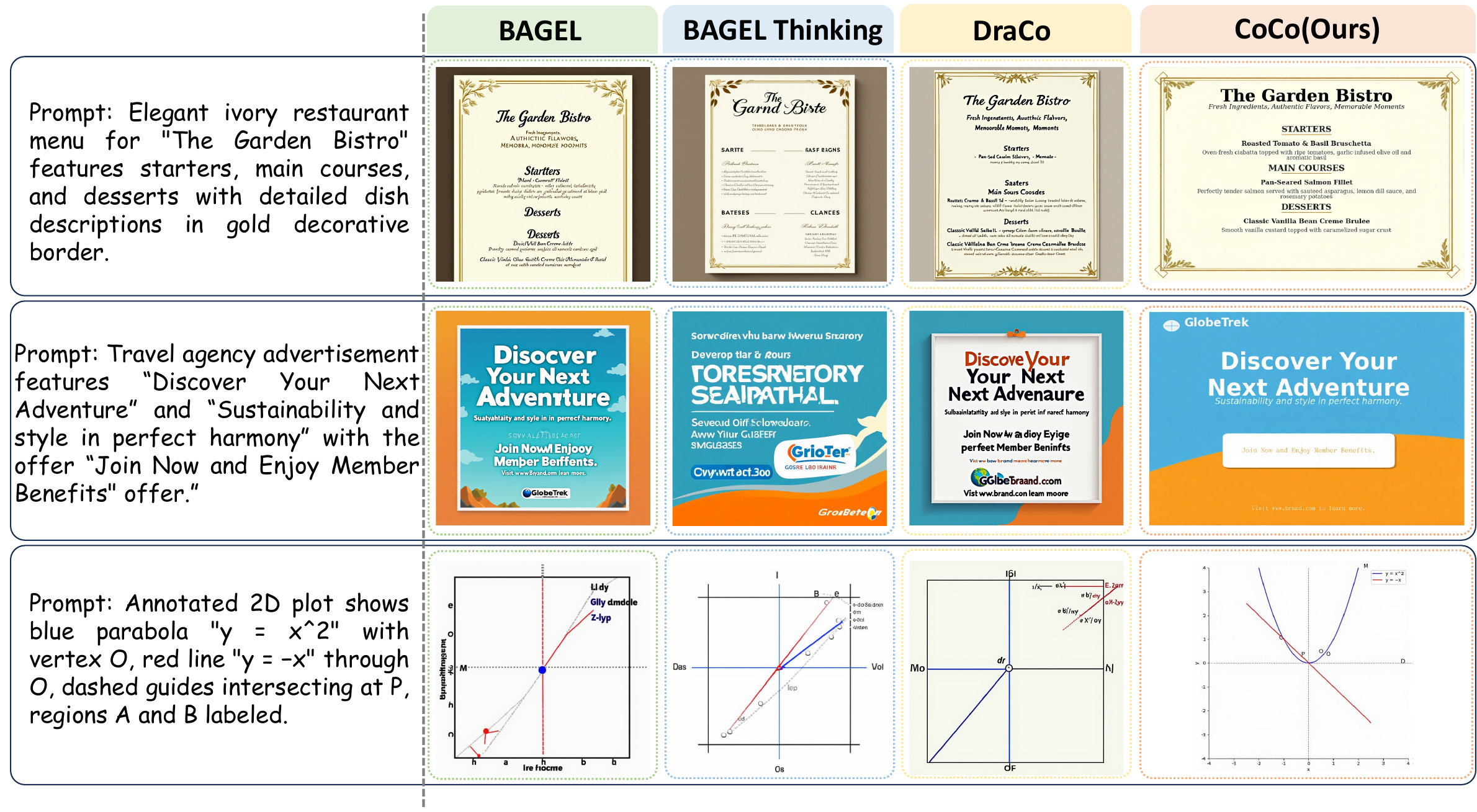}
    \caption{\textbf{Contrastive visualization of the experiment.}
    \coco{} generates accurate layouts and high-fidelity text, outperforming baselines,
    and supports adaptive aspect ratios that align images with prompt semantics.}
    \label{fig:vis1}
\end{figure*}

\paragraph{Evaluation.}
We evaluate our method on StructT2IBench~\citep{zhuo2025factualitymattersimagegeneration}, LongText-Bench~\citep{geng2025xomnireinforcementlearningmakes}, and OneIG-Bench~\citep{chang2025oneigbenchomnidimensionalnuancedevaluation}.
StructT2IBench evaluates image generation and editing on structured visuals, including charts, diagrams, mathematical figures, tables, and puzzles, and contains 1,714 evaluation samples with 37,941 Q\&A pairs.
For text rendering evaluation, we adopt the text rendering subset of OneIG-Bench, covering both English and Chinese tasks and using a composite metric combining Edit Distance, Completion Rate, and Word Accuracy to assess typographic precision.
We further evaluate long-form text generation on LongText-Bench, which contains 160 prompts across eight scenarios targeting accurate rendering of long Chinese and English text.
Following the official protocols, we generate one image per prompt for StructT2IBench and LongText-Bench, and four images per prompt for OneIG-Bench.

\paragraph{Training Details.}
We train our model using the {\bf \coco{}}-10K and StructVisuals~\citep{zhuo2025factualitymattersimagegeneration} datasets. To preserve visual perception ability, we also utilize the caption annotations from StructVisuals during training. We empirically observe that Bagel~\citep{deng2025emerging} struggles to generate executable code, so we first perform a text-to-code fine-tuning stage to equip the model with basic code generation capability. Afterward, we conduct full-parameter fine-tuning for 16K steps and adopt the EMA weights for evaluation. The learning rate is set to $2\times10^{-5}$ with 2K warmup steps. During training, we freeze the ViT encoder, the VAE encoder, and their connector to preserve stable visual encoding and reconstruction ability. The model is trained using 8 H800 GPUs.

\subsection{Main Results}

We present quantitative results in Tables~\ref{tab:struct_t2i} and \ref{tab:text_render}, with qualitative results shown in Fig.~\ref{fig:vis1}. We compare our method against generation-only models (including diffusion-based and autoregressive models), unified MLLMs, and unified MLLMs with CoT planning.

\begin{table*}[t]
    \centering
    \caption{\textbf{Quantitative comparison on StructT2IBench.} All values are reported as Accuracy (\%). \colorbox{goldcolor}{Orange} and \colorbox{silvercolor}{Champagne} cells indicate the Top 1 and Top 2 performers respectively.}
    \label{tab:struct_t2i}
    \small  
    \renewcommand{\arraystretch}{1.2}  
    \resizebox{1.0\linewidth}{!}{
        \begin{tabular}{l|cccccccc}
            \toprule
            Model & \textbf{Chart}$\uparrow$ & \textbf{Graph}$\uparrow$ & \textbf{Math}$\uparrow$ & \textbf{Puzzle}$\uparrow$ & \textbf{Science}$\uparrow$ & \textbf{Table}$\uparrow$ & \textbf{Overall}$\uparrow$ \\
            \midrule
            \multicolumn{5}{l}{\textit{\color{gray}Close-resource models}} \\
            Seedream 4.0 ~\cite{seedream2025seedream}& 35.79 & 54.08 & 63.33 & 50.89 & \cellcolor{goldcolor}62.59 & 68.94 & 47.52 \\
            Nano banana ~\cite{bananapro}& 35.55 & \cellcolor{silvercolor}58.96 & \cellcolor{silvercolor}64.81 & \cellcolor{goldcolor}63.87 & 60.75 & 67.20 & 48.45 \\
            GPT-Image ~\cite{openai2024gpt4o}& \cellcolor{silvercolor}37.09 & 57.00 & 63.25 & \cellcolor{silvercolor}59.42 & \cellcolor{silvercolor}60.94 & \cellcolor{goldcolor}83.31 & \cellcolor{silvercolor}49.58 \\
            \midrule
            \multicolumn{5}{l}{\textit{\color{gray}Open-resource models}} \\
            UniWorld-V1 ~\cite{lin2025uniworld}& 1.71 & 5.52 & 4.72 & 1.58 & 8.82 & 5.25 & 3.20 \\
            Bagel ~\cite{deng2025emerging}& 4.66 & 3.61 & 4.02 & 4.46 & 8.60 & 5.74 & 4.69 \\
            Bagel-Think ~\cite{deng2025emerging}& 4.81 & 15.33 & 13.89 & 15.22 & 19.05 & 8.97 & 9.03 \\
            HiDream-I1-Full ~\cite{cai2025hidream}& 9.47 & 20.84 & 19.20 & 18.00 & 26.77 & 27.05 & 14.77 \\
            OmniGen2 ~\cite{wu2025omnigen2}& 10.67 & 22.51 & 22.89 & 18.63 & 28.00 & 22.61 & 16.24 \\
            FLUX.1 Dev ~\cite{labs2025flux1kontextflowmatching}& 12.35 & 20.09 & 19.86 & 20.63 & 25.25 & 27.00 & 16.51 \\
            FLUX.1 Kontext ~\cite{labs2025flux1kontextflowmatching}& 17.22 & 24.64 & 21.42 & 24.06 & 30.97 & 29.16 & 20.36 \\
            Ovis-U1 ~\cite{wang2025ovis}& 24.75 & 16.08 & 19.45 & 21.23 & 26.03 & 12.70 & 22.83 \\
            Qwen-Image ~\cite{wu2025qwen}& 32.23 & 48.05 & 46.98 & 48.90 & 53.51 & 73.65 & 41.03 \\
            \midrule
            \textbf{\coco{} (Ours)} & \cellcolor{goldcolor}79.44 & \cellcolor{goldcolor}62.58 & \cellcolor{goldcolor}69.12 & 49.10 & 58.81 & \cellcolor{silvercolor}79.15 & \cellcolor{goldcolor}73.52 \\
            \bottomrule
        \end{tabular}
    }
\end{table*}

Our method substantially outperforms other methods on StructT2IBench, achieving an overall accuracy of 73.52\%, significantly surpassing the best baseline (GPT-Image, 49.58\%). Notably, \coco{} achieves the best performance on several structurally demanding tasks, including Chart (79.44\%), Graph (62.58\%), Math (69.12\%), and Table (79.15\%), which require precise spatial layouts and structured reasoning. These results demonstrate the effectiveness of code-based reasoning for handling complex structural compositions in text-to-image generation.

  \begin{table}[t]
      \centering
      \caption{Comparison of text rendering capability. All values are reported as Accuracy (\%). \colorbox{goldcolor}{Orange} and \colorbox{silvercolor}{Champagne} cells indicate the Top 1 and Top 2 performers respectively.}
      \label{tab:text_render}
      \small
      \setlength{\tabcolsep}{5pt}
      \renewcommand{\arraystretch}{1.08}
      
      \begin{tabular}{lcccccc}
          \toprule
          \multirow{2}{*}{Method} & \multicolumn{3}{c}{OneIG-Bench~\cite{chang2025oneigbenchomnidimensionalnuancedevaluation}} & \multicolumn{3}{c}{LongText-Bench~\cite{geng2025xomnireinforcementlearningmakes}} \\
          \cmidrule(lr){2-4} \cmidrule(lr){5-7}
          & English & Chinese & Overall & English & Chinese & Overall\\
          \midrule
          \multicolumn{7}{l}{\textit{\color{gray}Gen. Only Models}} \\
          FLUX.1-dev~\cite{labs2025flux1kontextflowmatching}     & 0.523 & -  & -   & 0.607 & 0.005 & 0.306\\
          HiDream-I1~\cite{cai2025hidream} & 0.707 & 0.205 & 0.456 & 0.543 & 0.024 & 0.284\\
          Kolors 2.0~\cite{team2025okuaishou}     & 0.427 & 0.502 & 0.465 & 0.258 & 0.329 & 0.294\\
          \midrule
          \multicolumn{7}{l}{\textit{\color{gray}Unified Models}} \\
          Janus-Pro~\cite{chen2025janus}      & 0.019 & 0.015 & 0.017 & 0.019 & 0.006 & 0.013\\
          BLIP3-o~\cite{chen2025blip3}       & 0.013 & 0.092 & 0.053 & 0.021 & 0.018 & 0.020\\
          
          OmniGen2~\cite{xiao2025omnigen}       & 0.680 & -    & -  & 0.561 & 0.059 & 0.310 \\
          Show-o2~\cite{xie2025show}       & 0.002 & -   & -   & 0.006 & 0.002 & 0.004\\
          BAGEL~\cite{deng2025emerging}         & 0.244 & 0.365 & 0.305 & 0.373 & 0.310 & 0.342\\
          
          GPT-4o~\cite{openai2024gpt4o}         & \cellcolor{silvercolor}0.857 & \cellcolor{silvercolor}0.650 & \cellcolor{silvercolor}0.754 & \cellcolor{goldcolor}0.956 & \cellcolor{silvercolor}0.619 & \cellcolor{goldcolor}0.788 \\
          \midrule
          \multicolumn{7}{l}{\textit{\color{gray}Unified Model w/ CoT}} \\
          BAGEL-thinking~\cite{deng2025emerging}  & 0.020 & 0.127 & 0.074 & 0.068 & 0.105 & 0.087 \\
          \textbf{CoCo(Ours)}     & \cellcolor{goldcolor}0.895 & \cellcolor{goldcolor}0.811 & \cellcolor{goldcolor}0.853 & \cellcolor{silvercolor}0.755 & \cellcolor{goldcolor}0.753 & \cellcolor{silvercolor}0.754 \\
          \bottomrule
      \end{tabular}
  \end{table}

Furthermore, on text rendering benchmarks, \coco{} achieves strong performance across both datasets. On OneIG-Bench, it reaches 0.895 (English) and 0.811 (Chinese) with an overall score of 0.853, outperforming all compared generation-only and unified MLLM methods. On LongText-Bench, \coco{} achieves 0.755 (English) and 0.753 (Chinese) with an overall score of 0.754, demonstrating competitive performance on long-text rendering. These results validate that code-based intermediate representations enable more reliable handling of complex textual instructions and structured visual layouts.

\subsection{Ablation Study}

To better understand the contribution of each component, we conduct ablation studies on the two key learnable modules in CoCo: code generation and draft-guided refinement.
  \begin{figure*}[tp]
      \centering
      \begin{minipage}{0.5\textwidth}
          \centering
          \renewcommand{\arraystretch}{1.4}
          \small
          \begin{tabular}{@{}lccc@{}}
              \toprule
              Method & $r_c$ & English & Chinese \\
              \midrule
              Bagel & -- & 0.373 & 0.310 \\
              \midrule
              CoCo & 0.20 & 0.724 & 0.667 \\
              CoCo & 0.10 & 0.733 & 0.671 \\
              CoCo & \textbf{0.05} & \textbf{0.755} & \textbf{0.753} \\
              \bottomrule
          \end{tabular}
          \setlength{\abovecaptionskip}{15pt}
          \captionof{table}{\textbf{Ablation on training mixture ratio.} $r_c$ is the proportion of Text--Code supervision.}
          \label{tab:ablation}
      \end{minipage}
      \hfill
      \begin{minipage}{0.45\textwidth}
          \centering
          \includegraphics[width=0.9\linewidth]{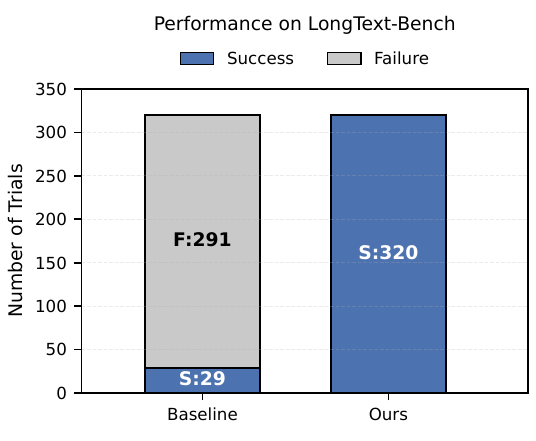}
          \setlength{\abovecaptionskip}{-5pt}
          \captionof{figure}{\textbf{Code executability diagnostic.} CoCo vs. Bagel on LongText-Bench.}
          \label{fig:comparison}
      \end{minipage}
  \end{figure*}

\paragraph{\textbf{Effect of Training Mixture Ratios.}}
Table~\ref{tab:ablation} examines the effect of different training-mixture ratios on model performance. Following the training strategy described in Sec.~\ref{dataset: training}, we train the model with a mixture of two complementary supervision sources: (1) Text--Code pairs and (2) Text-Draft Image-Final Image triplets. The former promotes executable reasoning, while the latter teaches the model to refine coarse draft renderings into final images with higher visual fidelity and stronger semantic alignment with the input text. As the proportion of Text--Code supervision decreases, performance improves consistently on both the English and Chinese benchmarks. The best results are obtained at $r_c=0.05$, where CoCo achieves 0.755 on EN and 0.753 on CN, substantially outperforming the Bagel baseline. These results indicate that only a small amount of code supervision is needed to induce structured reasoning, whereas the dominant training signal should come from draft-to-final refinement data to better support faithful and semantically accurate rendering.

\paragraph{\textbf{Is Text-Code Supervision Necessary?}}
We further evaluate the role of Text--Code supervision in structured generation. As shown in Fig.~\ref{fig:comparison}, the off-the-shelf Bagel~\citep{deng2025emerging}, without training on Text--Code pairs, often produces non-executable code, leading to a high rendering failure rate. On LongText-Bench, only 9.06\% of its generated programs compile successfully (29/320), indicating that the model struggles to produce reliable code for downstream rendering. In contrast, after fine-tuning with our training pipeline, \coco{} achieves a 100\% compilation success rate. This result shows that Text--Code supervision is essential for teaching the model to generate executable code, which in turn serves as a prerequisite for stable and successful rendering.

\subsection{Discussion and Further Analysis}

We observe an interesting generalization behavior of \coco{}. Although all training data are constructed at a fixed resolution of $1024$, the generated code during inference is not restricted to this setting. Instead, the model adaptively configures the canvas according to the prompt semantics. For example, poster-like prompts often lead to wider layouts (e.g., $16{:}9$), while charts or diagrams typically produce square or near-square canvases. Qualitative examples in Fig.~\ref{fig:vis1} show that \coco{} naturally generates outputs with different aspect ratios depending on the prompt.

This behavior suggests that the model does not simply memorize a fixed rendering configuration. Instead, it learns to treat canvas size and layout parameters as part of the reasoning process encoded in the generated code. We attribute this flexibility to the programmatic nature of Code-as-CoT, where spatial layouts and canvas parameters are explicitly represented in code and can be dynamically adjusted to match the prompt.

\section{Conclusion}

In this work, we introduce \textbf{\coco{}}, a novel code-driven reasoning framework that formulates executable code as Chain-of-Thought for text-to-image generation. Instead of relying on abstract natural-language planning, \coco{} generates structured code that explicitly specifies spatial layouts and textual elements, which is executed to produce an intermediate draft image. The model then performs fine-grained image editing on the draft to obtain the final high-fidelity result. To support this paradigm, we build \coco{}-10K with Text-Code pairs and Text-Draft Image-Final Image triplets, where draft-to-final supervision yields higher-fidelity outputs with stronger semantic alignment. Extensive experiments demonstrate the effectiveness of our approach, showing substantial improvements on challenging benchmarks including StructT2IBench, OneIG-Bench, and LongText-Bench. Our results highlight executable code as a reliable intermediate reasoning representation for precise, controllable, and structured text-to-image generation.

\newpage
\section*{Acknowledgments}

This work is partially supported by the following grants: Guangdong Provincial Key Fields Research and Development Program (2025B0101120004), National Natural Science Foundation of China (61972163), Natural Science Foundation of Guangdong Province (2023A1515012568), Guangdong Provincial Key Laboratory of Human Digital Twin (2022B1212010004).

\bibliographystyle{unsrtnat}
\bibliography{main}

@String(ECCV= {Eur. Conf. Comput. Vis.})

@String(ECCV  = {ECCV})

@article{zhou2024transfusion,
  title={Transfusion: Predict the next token and diffuse images with one multi-modal model},
  author={Zhou, Chunting and Yu, Lili and Babu, Arun and Tirumala, Kushal and Yasunaga, Michihiro and Shamis, Leonid and Kahn, Jacob and Ma, Xuezhe and Zettlemoyer, Luke and Levy, Omer},
  journal={arXiv preprint arXiv:2408.11039},
  year={2024}
}

@article{liquid,
  title={Liquid: Language models are scalable multi-modal generators},
  author={Wu, Junfeng and Jiang, Yi and Ma, Chuofan and Liu, Yuliang and Zhao, Hengshuang and Yuan, Zehuan and Bai, Song and Bai, Xiang},
  journal={arXiv preprint arXiv:2412.04332},
  year={2024}
}

@article{wang2024emu3,
  title={Emu3: Next-token prediction is all you need},
  author={Wang, Xinlong and Zhang, Xiaosong and Luo, Zhengxiong and Sun, Quan and Cui, Yufeng and Wang, Jinsheng and Zhang, Fan and Wang, Yueze and Li, Zhen and Yu, Qiying and others},
  journal={arXiv preprint arXiv:2409.18869},
  year={2024}
}

@article{qu2024tokenflow,
  title={Tokenflow: Unified image tokenizer for multimodal understanding and generation},
  author={Qu, Liao and Zhang, Huichao and Liu, Yiheng and Wang, Xu and Jiang, Yi and Gao, Yiming and Ye, Hu and Du, Daniel K and Yuan, Zehuan and Wu, Xinglong},
  journal={arXiv preprint arXiv:2412.03069},
  year={2024}
}

@article{team2024chameleon,
  title={Chameleon: Mixed-modal early-fusion foundation models},
  author={Team, Chameleon},
  journal={arXiv preprint arXiv:2405.09818},
  year={2024}
}

@article{guo2025deepseek,
  title={DeepSeek-R1: Incentivizing Reasoning Capability in LLMs via Reinforcement Learning},
  author={Guo, Daya and Yang, Dejian and Zhang, Haowei and Song, Junxiao and Zhang, Ruoyu and Xu, Runxin and Zhu, Qihao and Ma, Shirong and Wang, Peiyi and Bi, Xiao and others},
  journal={arXiv preprint arXiv:2501.12948},
  year={2025}
}

@article{wei2022chain,
  title={Chain-of-thought prompting elicits reasoning in large language models},
  author={Wei, Jason and Wang, Xuezhi and Schuurmans, Dale and Bosma, Maarten and Xia, Fei and Chi, Ed and Le, Quoc V and Zhou, Denny and others},
  journal={Advances in neural information processing systems},
  volume={35},
  pages={24824--24837},
  year={2022}
}

@article{jiang2025mme,
  title={MME-CoT: Benchmarking Chain-of-Thought in Large Multimodal Models for Reasoning Quality, Robustness, and Efficiency},
  author={Jiang, Dongzhi and Zhang, Renrui and Guo, Ziyu and Li, Yanwei and Qi, Yu and Chen, Xinyan and Wang, Liuhui and Jin, Jianhan and Guo, Claire and Yan, Shen and others},
  journal={arXiv preprint arXiv:2502.09621},
  year={2025}
}

@misc{chen2025r1v,
  author       = {Chen, Liang and Li, Lei and Zhao, Haozhe and Song, Yifan and Vinci},
  title        = {R1-V: Reinforcing Super Generalization Ability in Vision-Language Models with Less Than \$3},
  howpublished = {\url{https://github.com/Deep-Agent/R1-V}},
  note         = {Accessed: 2025-02-02},
  year         = {2025}
}

@article{zhan2025visionr1,
  title={Vision-R1: Evolving Human-Free Alignment in Large Vision-Language Models via Vision-Guided Reinforcement Learning},
  author={Zhan, Yufei and Zhu, Yousong and Zheng, Shurong and Zhao, Hongyin and Yang, Fan and Tang, Ming and Wang, Jinqiao},
  journal={arXiv preprint arXiv:2503.18013},
  year={2025}
}

@article{wei2025webagent,
  title={Webagent-r1: Training web agents via end-to-end multi-turn reinforcement learning},
  author={Wei, Zhepei and Yao, Wenlin and Liu, Yao and Zhang, Weizhi and Lu, Qin and Qiu, Liang and Yu, Changlong and Xu, Puyang and Zhang, Chao and Yin, Bing and others},
  journal={arXiv preprint arXiv:2505.16421},
  year={2025}
}

@article{mai2025agent,
  title={Agent rl scaling law: Agent rl with spontaneous code execution for mathematical problem solving},
  author={Mai, Xinji and Xu, Haotian and Li, Zhong-Zhi and Wang, Weinong and Hu, Jian and Zhang, Yingying and Zhang, Wenqiang and others},
  journal={arXiv preprint arXiv:2505.07773},
  year={2025}
}

@article{zhang2024mathverse,
  title={Mathverse: Does your multi-modal llm truly see the diagrams in visual math problems?},
  author={Zhang, Renrui and Jiang, Dongzhi and Zhang, Yichi and Lin, Haokun and Guo, Ziyu and Qiu, Pengshuo and Zhou, Aojun and Lu, Pan and Chang, Kai-Wei and Gao, Peng and others},
  journal={ECCV 2024},
  year={2024}
}

@article{Lu2023MathVistaEM,
  title={MathVista: Evaluating Math Reasoning in Visual Contexts with GPT-4V, Bard, and Other Large Multimodal Models},
  author={Pan Lu and Hritik Bansal and Tony Xia and Jiacheng Liu and Chun-yue Li and Hannaneh Hajishirzi and Hao Cheng and Kai-Wei Chang and Michel Galley and Jianfeng Gao},
  journal={ArXiv},
  year={2023},
  volume={abs/2310.02255},
}

@article{gu2025improving,
  title={Improving Chain-of-Thought Efficiency for Autoregressive Image Generation},
  author={Gu, Zeqi and Georgopoulos, Markos and Dai, Xiaoliang and Ghazvininejad, Marjan and Wang, Chu and Juefei-Xu, Felix and Li, Kunpeng and Shi, Yujun and He, Zecheng and He, Zijian and others},
  journal={arXiv preprint arXiv:2510.05593},
  year={2025}
}

@article{wu2024janus,
  title={Janus: Decoupling visual encoding for unified multimodal understanding and generation},
  author={Wu, Chengyue and Chen, Xiaokang and Wu, Zhiyu and Ma, Yiyang and Liu, Xingchao and Pan, Zizheng and Liu, Wen and Xie, Zhenda and Yu, Xingkai and Ruan, Chong and others},
  journal={arXiv preprint arXiv:2410.13848},
  year={2024}
}

@article{cai2025hidream,
  title={Hidream-i1: A high-efficient image generative foundation model with sparse diffusion transformer},
  author={Cai, Qi and Chen, Jingwen and Chen, Yang and Li, Yehao and Long, Fuchen and Pan, Yingwei and Qiu, Zhaofan and Zhang, Yiheng and Gao, Fengbin and Xu, Peihan and others},
  journal={arXiv preprint arXiv:2505.22705},
  year={2025}
}

@inproceedings{nguyen2025yo,
  title={Yo'Chameleon: Personalized Vision and Language Generation},
  author={Nguyen, Thao and Singh, Krishna Kumar and Shi, Jing and Bui, Trung and Lee, Yong Jae and Li, Yuheng},
  booktitle={Proceedings of the Computer Vision and Pattern Recognition Conference},
  pages={14438--14448},
  year={2025}
}

@article{zhong2026unified,
  title={Unified Personalized Understanding, Generating and Editing},
  author={Zhong, Yu and Lin, Tianwei and Zhu, Ruike and Yuan, Yuqian and Zheng, Haoyu and Liang, Liang and Zhang, Wenqiao and Shao, Feifei and Li, Haoyuan and He, Wanggui and others},
  journal={arXiv preprint arXiv:2601.06965},
  year={2026}
}

@article{an2025unictokens,
  title={Unictokens: Boosting personalized understanding and generation via unified concept tokens},
  author={An, Ruichuan and Yang, Sihan and Zhang, Renrui and Shen, Zijun and Lu, Ming and Dai, Gaole and Liang, Hao and Guo, Ziyu and Yan, Shilin and Luo, Yulin and others},
  journal={arXiv preprint arXiv:2505.14671},
  year={2025}
}

@inproceedings{xiao2025omnigen,
  title={Omnigen: Unified image generation},
  author={Xiao, Shitao and Wang, Yueze and Zhou, Junjie and Yuan, Huaying and Xing, Xingrun and Yan, Ruiran and Li, Chaofan and Wang, Shuting and Huang, Tiejun and Liu, Zheng},
  booktitle={Proceedings of the Computer Vision and Pattern Recognition Conference},
  pages={13294--13304},
  year={2025}
}

@article{wu2024vila,
  title={Vila-u: a unified foundation model integrating visual understanding and generation},
  author={Wu, Yecheng and Zhang, Zhuoyang and Chen, Junyu and Tang, Haotian and Li, Dacheng and Fang, Yunhao and Zhu, Ligeng and Xie, Enze and Yin, Hongxu and Yi, Li and others},
  journal={arXiv preprint arXiv:2409.04429},
  year={2024}
}

@article{huang2025interleaving,
  title={Interleaving reasoning for better text-to-image generation},
  author={Huang, Wenxuan and Chen, Shuang and Xie, Zheyong and Cao, Shaosheng and Tang, Shixiang and Shen, Yufan and Yin, Qingyu and Hu, Wenbo and Wang, Xiaoman and Tang, Yuntian and others},
  journal={arXiv preprint arXiv:2509.06945},
  year={2025}
}

@article{zhang2025scaling,
  title={Scaling and Beyond: Advancing Spatial Reasoning in MLLMs Requires New Recipes},
  author={Zhang, Huanyu and Li, Chengzu and Wu, Wenshan and Mao, Shaoguang and Zhang, Yifan and Tian, Haochen and Vuli{\'c}, Ivan and Zhang, Zhang and Wang, Liang and Tan, Tieniu and others},
  journal={arXiv preprint arXiv:2504.15037},
  year={2025}
}

@misc{bananapro,
  title = {Introducing nano banana pro},
  author = {{Google}},
  year = {2025},
  month = {November},
  howpublished = {\url{https://blog.google/technology/ai/nano-banana-pro/}},
}

@misc{gemini,
  title = {A new era of intelligence with Gemini 3},
  author = {{Google}},
  year = {2025},
  month = {November},
  howpublished = {\url{https://blog.google/products-and-platforms/products/gemini/gemini-3/}},
}

@article{zhang2025latent,
  title={Latent Sketchpad: Sketching Visual Thoughts to Elicit Multimodal Reasoning in MLLMs},
  author={Zhang, Huanyu and Wu, Wenshan and Li, Chengzu and Shang, Ning and Xia, Yan and Huang, Yangyu and Zhang, Yifan and Dong, Li and Zhang, Zhang and Wang, Liang and others},
  journal={arXiv preprint arXiv:2510.24514},
  year={2025}
}

@article{yue2023mmmu,
  title={MMMU: A Massive Multi-discipline Multimodal Understanding and Reasoning Benchmark for Expert AGI},
  author={Xiang Yue and Yuansheng Ni and Kai Zhang and Tianyu Zheng and Ruoqi Liu and Ge Zhang and Samuel Stevens and Dongfu Jiang and Weiming Ren and Yuxuan Sun and Cong Wei and Botao Yu and Ruibin Yuan and Renliang Sun and Ming Yin and Boyuan Zheng and Zhenzhu Yang and Yibo Liu and Wenhao Huang and Huan Sun and Yu Su and Wenhu Chen},
  journal={arXiv preprint arXiv:2311.16502},
  year={2023},
}

@article{zhang2025reasongen,
  title={ReasonGen-R1: CoT for Autoregressive Image generation models through SFT and RL},
  author={Zhang, Yu and Li, Yunqi and Yang, Yifan and Wang, Rui and Yang, Yuqing and Qi, Dai and Bao, Jianmin and Chen, Dongdong and Luo, Chong and Qiu, Lili},
  journal={arXiv preprint arXiv:2505.24875},
  year={2025}
}

@article{pan2025focusdiff,
  title={FocusDiff: Advancing Fine-Grained Text-Image Alignment for Autoregressive Visual Generation through RL},
  author={Pan, Kaihang and Bu, Wendong and Wu, Yuruo and Wu, Yang and Shen, Kai and Li, Yunfei and Zhao, Hang and Li, Juncheng and Tang, Siliang and Zhuang, Yueting},
  journal={arXiv preprint arXiv:2506.05501},
  year={2025}
}

@article{wu2025qwen,
  title={Qwen-image technical report},
  author={Wu, Chenfei and Li, Jiahao and Zhou, Jingren and Lin, Junyang and Gao, Kaiyuan and Yan, Kun and Yin, Sheng-ming and Bai, Shuai and Xu, Xiao and Chen, Yilei and others},
  journal={arXiv preprint arXiv:2508.02324},
  year={2025}
}

@article{wu2025omnigen2,
  title={OmniGen2: Exploration to Advanced Multimodal Generation},
  author={Wu, Chenyuan and Zheng, Pengfei and Yan, Ruiran and Xiao, Shitao and Luo, Xin and Wang, Yueze and Li, Wanli and Jiang, Xiyan and Liu, Yexin and Zhou, Junjie and others},
  journal={arXiv preprint arXiv:2506.18871},
  year={2025}
}

@article{lin2025uniworld,
  title={Uniworld: High-resolution semantic encoders for unified visual understanding and generation},
  author={Lin, Bin and Li, Zongjian and Cheng, Xinhua and Niu, Yuwei and Ye, Yang and He, Xianyi and Yuan, Shenghai and Yu, Wangbo and Wang, Shaodong and Ge, Yunyang and others},
  journal={arXiv preprint arXiv:2506.03147},
  year={2025}
}

@article{xin2025lumina,
  title={Lumina-dimoo: An omni diffusion large language model for multi-modal generation and understanding},
  author={Xin, Yi and Qin, Qi and Luo, Siqi and Zhu, Kaiwen and Yan, Juncheng and Tai, Yan and Lei, Jiayi and Cao, Yuewen and Wang, Keqi and Wang, Yibin and others},
  journal={arXiv preprint arXiv:2510.06308},
  year={2025}
}

@article{kojima2022large,
  title={Large language models are zero-shot reasoners},
  author={Kojima, Takeshi and Gu, Shixiang Shane and Reid, Machel and Matsuo, Yutaka and Iwasawa, Yusuke},
  journal={Advances in neural information processing systems},
  volume={35},
  pages={22199--22213},
  year={2022}
}

@article{meng2025mm,
  title={MM-Eureka: Exploring Visual Aha Moment with Rule-based Large-scale Reinforcement Learning},
  author={Meng, Fanqing and Du, Lingxiao and Liu, Zongkai and Zhou, Zhixiang and Lu, Quanfeng and Fu, Daocheng and Shi, Botian and Wang, Wenhai and He, Junjun and Zhang, Kaipeng and others},
  journal={arXiv preprint arXiv:2503.07365},
  year={2025}
}

@article{amini2019mathqa,
  title={Mathqa: Towards interpretable math word problem solving with operation-based formalisms},
  author={Amini, Aida and Gabriel, Saadia and Lin, Peter and Koncel-Kedziorski, Rik and Choi, Yejin and Hajishirzi, Hannaneh},
  journal={arXiv preprint arXiv:1905.13319},
  year={2019}
}

@article{hendrycksmath2021,
  title={Measuring Mathematical Problem Solving With the MATH Dataset},
  author={Dan Hendrycks and Collin Burns and Saurav Kadavath and Akul Arora and Steven Basart and Eric Tang and Dawn Song and Jacob Steinhardt},
  journal={NeurIPS},
  year={2021}
}

@article{tschannen2025siglip,
  title={Siglip 2: Multilingual vision-language encoders with improved semantic understanding, localization, and dense features},
  author={Tschannen, Michael and Gritsenko, Alexey and Wang, Xiao and Naeem, Muhammad Ferjad and Alabdulmohsin, Ibrahim and Parthasarathy, Nikhil and Evans, Talfan and Beyer, Lucas and Xia, Ye and Mustafa, Basil and others},
  journal={arXiv preprint arXiv:2502.14786},
  year={2025}
}

@article{kingma2013auto,
  title={Auto-encoding variational bayes},
  author={Kingma, Diederik P and Welling, Max},
  journal={arXiv preprint arXiv:1312.6114},
  year={2013}
}

@article{vaswani2017attention,
  title={Attention is all you need},
  author={Vaswani, Ashish and Shazeer, Noam and Parmar, Niki and Uszkoreit, Jakob and Jones, Llion and Gomez, Aidan N and Kaiser, {\L}ukasz and Polosukhin, Illia},
  journal={Advances in neural information processing systems},
  volume={30},
  year={2017}
}

@inproceedings{esser2024scaling,
  title={Scaling rectified flow transformers for high-resolution image synthesis},
  author={Esser, Patrick and Kulal, Sumith and Blattmann, Andreas and Entezari, Rahim and M{\"u}ller, Jonas and Saini, Harry and Levi, Yam and Lorenz, Dominik and Sauer, Axel and Boesel, Frederic and others},
  booktitle={Forty-first international conference on machine learning},
  year={2024}
}

@article{lipman2022flow,
  title={Flow matching for generative modeling},
  author={Lipman, Yaron and Chen, Ricky TQ and Ben-Hamu, Heli and Nickel, Maximilian and Le, Matt},
  journal={arXiv preprint arXiv:2210.02747},
  year={2022}
}

@article{liu2022flow,
  title={Flow straight and fast: Learning to generate and transfer data with rectified flow},
  author={Liu, Xingchao and Gong, Chengyue and Liu, Qiang},
  journal={arXiv preprint arXiv:2209.03003},
  year={2022}
}

@misc{openai2024gpt4o,
  author    = {OpenAI},
  title     = {Hello GPT-4o},
  howpublished = {\url{https://openai.com/index/hello-gpt-4o/}},
  year      = {2024}
}

@misc{labs2025flux1kontextflowmatching,
      title={FLUX.1 Kontext: Flow Matching for In-Context Image Generation and Editing in Latent Space},
      author={Black Forest Labs and Stephen Batifol and Andreas Blattmann and Frederic Boesel and Saksham Consul and Cyril Diagne and Tim Dockhorn and Jack English and Zion English and Patrick Esser and Sumith Kulal and Kyle Lacey and Yam Levi and Cheng Li and Dominik Lorenz and Jonas Müller and Dustin Podell and Robin Rombach and Harry Saini and Axel Sauer and Luke Smith},
      year={2025},
      eprint={2506.15742},
      archivePrefix={arXiv},
      primaryClass={cs.GR},
      url={https://arxiv.org/abs/2506.15742},
}

@article{wang2025ovis,
  title={Ovis-u1 technical report},
  author={Wang, Guo-Hua and Zhao, Shanshan and Zhang, Xinjie and Cao, Liangfu and Zhan, Pengxin and Duan, Lunhao and Lu, Shiyin and Fu, Minghao and Chen, Xiaohao and Zhao, Jianshan and others},
  journal={arXiv preprint arXiv:2506.23044},
  year={2025}
}

@article{cui2025emu3,
  title={Emu3. 5: Native multimodal models are world learners},
  author={Cui, Yufeng and Chen, Honghao and Deng, Haoge and Huang, Xu and Li, Xinghang and Liu, Jirong and Liu, Yang and Luo, Zhuoyan and Wang, Jinsheng and Wang, Wenxuan and others},
  journal={arXiv preprint arXiv:2510.26583},
  year={2025}
}

@inproceedings{tong2025metamorph,
  title={Metamorph: Multimodal understanding and generation via instruction tuning},
  author={Tong, Shengbang and Fan, David and Li, Jiachen and Xiong, Yunyang and Chen, Xinlei and Sinha, Koustuv and Rabbat, Michael and LeCun, Yann and Xie, Saining and Liu, Zhuang},
  booktitle={Proceedings of the IEEE/CVF International Conference on Computer Vision},
  pages={17001--17012},
  year={2025}
}

@article{xie2024show,
  title={Show-o: One single transformer to unify multimodal understanding and generation},
  author={Xie, Jinheng and Mao, Weijia and Bai, Zechen and Zhang, David Junhao and Wang, Weihao and Lin, Kevin Qinghong and Gu, Yuchao and Chen, Zhijie and Yang, Zhenheng and Shou, Mike Zheng},
  journal={arXiv preprint arXiv:2408.12528},
  year={2024}
}

@article{zhao2024monoformer,
  title={Monoformer: One transformer for both diffusion and autoregression},
  author={Zhao, Chuyang and Song, Yuxing and Wang, Wenhao and Feng, Haocheng and Ding, Errui and Sun, Yifan and Xiao, Xinyan and Wang, Jingdong},
  journal={arXiv preprint arXiv:2409.16280},
  year={2024}
}

@article{shi2024lmfusion,
  title={Lmfusion: Adapting pretrained language models for multimodal generation},
  author={Shi, Weijia and Han, Xiaochuang and Zhou, Chunting and Liang, Weixin and Lin, Xi Victoria and Zettlemoyer, Luke and Yu, Lili},
  journal={arXiv preprint arXiv:2412.15188},
  year={2024}
}

@article{liang2024mixture,
  title={Mixture-of-transformers: A sparse and scalable architecture for multi-modal foundation models},
  author={Liang, Weixin and Yu, Lili and Luo, Liang and Iyer, Srinivasan and Dong, Ning and Zhou, Chunting and Ghosh, Gargi and Lewis, Mike and Yih, Wen-tau and Zettlemoyer, Luke and others},
  journal={arXiv preprint arXiv:2411.04996},
  year={2024}
}

@article{chen2024diffusion,
  title={Diffusion forcing: Next-token prediction meets full-sequence diffusion},
  author={Chen, Boyuan and Mart{\'\i} Mons{\'o}, Diego and Du, Yilun and Simchowitz, Max and Tedrake, Russ and Sitzmann, Vincent},
  journal={Advances in Neural Information Processing Systems},
  volume={37},
  pages={24081--24125},
  year={2024}
}

@article{yu2025guided,
  title={Guided self-evolving llms with minimal human supervision},
  author={Yu, Wenhao and Liang, Zhenwen and Huang, Chengsong and Panaganti, Kishan and Fang, Tianqing and Mi, Haitao and Yu, Dong},
  journal={arXiv preprint arXiv:2512.02472},
  year={2025}
}

@article{zhou2024calibrated,
  title={Calibrated self-rewarding vision language models},
  author={Zhou, Yiyang and Fan, Zhiyuan and Cheng, Dongjie and Yang, Sihan and Chen, Zhaorun and Cui, Chenhang and Wang, Xiyao and Li, Yun and Zhang, Linjun and Yao, Huaxiu},
  journal={Advances in Neural Information Processing Systems},
  volume={37},
  pages={51503--51531},
  year={2024}
}

@article{wang2025unified,
  title={Unified reward model for multimodal understanding and generation},
  author={Wang, Yibin and Zang, Yuhang and Li, Hao and Jin, Cheng and Wang, Jiaqi},
  journal={arXiv preprint arXiv:2503.05236},
  year={2025}
}

@article{jin2025srum,
  title={Srum: Fine-grained self-rewarding for unified multimodal models},
  author={Jin, Weiyang and Niu, Yuwei and Liao, Jiaqi and Duan, Chengqi and Li, Aoxue and Gao, Shenghua and Liu, Xihui},
  journal={arXiv preprint arXiv:2510.12784},
  year={2025}
}

@misc{mao2025unirlselfimprovingunifiedmultimodal,
      title={UniRL: Self-Improving Unified Multimodal Models via Supervised and Reinforcement Learning},
      author={Weijia Mao and Zhenheng Yang and Mike Zheng Shou},
      year={2025},
      eprint={2505.23380},
      archivePrefix={arXiv},
      primaryClass={cs.CV},
      url={https://arxiv.org/abs/2505.23380},
}

@article{luo2025ursa,
  title={Ursa: Understanding and verifying chain-of-thought reasoning in multimodal mathematics},
  author={Luo, Ruilin and Zheng, Zhuofan and Wang, Yifan and Yu, Yiyao and Ni, Xinzhe and Lin, Zicheng and Zeng, Jin and Yang, Yujiu},
  journal={arXiv e-prints},
  pages={arXiv--2501},
  year={2025}
}

@article{peng2025lmm,
  title={Lmm-r1: Empowering 3b lmms with strong reasoning abilities through two-stage rule-based rl},
  author={Peng, Yingzhe and Zhang, Gongrui and Zhang, Miaosen and You, Zhiyuan and Liu, Jie and Zhu, Qipeng and Yang, Kai and Xu, Xingzhong and Geng, Xin and Yang, Xu},
  journal={arXiv preprint arXiv:2503.07536},
  year={2025}
}

@article{shen2025vlm,
  title={Vlm-r1: A stable and generalizable r1-style large vision-language model},
  author={Shen, Haozhan and Liu, Peng and Li, Jingcheng and Fang, Chunxin and Ma, Yibo and Liao, Jiajia and Shen, Qiaoli and Zhang, Zilun and Zhao, Kangjia and Zhang, Qianqian and others},
  journal={arXiv preprint arXiv:2504.07615},
  year={2025}
}

@article{tong2025delving,
  title={Delving into rl for image generation with cot: A study on dpo vs. grpo},
  author={Tong, Chengzhuo and Guo, Ziyu and Zhang, Renrui and Shan, Wenyu and Wei, Xinyu and Xing, Zhenghao and Li, Hongsheng and Heng, Pheng-Ann},
  journal={arXiv preprint arXiv:2505.17017},
  year={2025}
}

@article{wu2025visualquality,
  title={Visualquality-r1: Reasoning-induced image quality assessment via reinforcement learning to rank},
  author={Wu, Tianhe and Zou, Jian and Liang, Jie and Zhang, Lei and Ma, Kede},
  journal={arXiv preprint arXiv:2505.14460},
  year={2025}
}

@article{zhang2025perl,
  title={Perl: Permutation-enhanced reinforcement learning for interleaved vision-language reasoning},
  author={Zhang, Yizhen and Ding, Yang and Zhang, Shuoshuo and Zhang, Xinchen and Li, Haoling and Li, Zhong-zhi and Wang, Peijie and Wu, Jie and Ji, Lei and Shen, Yelong and others},
  journal={arXiv preprint arXiv:2506.14907},
  year={2025}
}

@article{chen2025blip3,
  title={Blip3-o: A family of fully open unified multimodal models-architecture, training and dataset},
  author={Chen, Jiuhai and Xu, Zhiyang and Pan, Xichen and Hu, Yushi and Qin, Can and Goldstein, Tom and Huang, Lifu and Zhou, Tianyi and Xie, Saining and Savarese, Silvio and others},
  journal={arXiv preprint arXiv:2505.09568},
  year={2025}
}

@article{guo2025can,
  title={Can We Generate Images with CoT? Let's Verify and Reinforce Image Generation Step by Step},
  author={Guo, Ziyu and Zhang, Renrui and Tong, Chengzhuo and Zhao, Zhizheng and Huang, Rui and Zhang, Haoquan and Zhang, Manyuan and Liu, Jiaming and Zhang, Shanghang and Gao, Peng and others},
  journal={arXiv preprint arXiv:2501.13926},
  year={2025}
}

@article{xie2025show,
  title={Show-o2: Improved native unified multimodal models},
  author={Xie, Jinheng and Yang, Zhenheng and Shou, Mike Zheng},
  journal={arXiv preprint arXiv:2506.15564},
  year={2025}
}

@article{yang2025hermesflow,
  title={Hermesflow: Seamlessly closing the gap in multimodal understanding and generation},
  author={Yang, Ling and Zhang, Xinchen and Tian, Ye and Shang, Chenming and Xu, Minghao and Zhang, Wentao and Cui, Bin},
  journal={arXiv preprint arXiv:2502.12148},
  year={2025}
}

@article{guo2025seed1,
  title={Seed1. 5-vl technical report},
  author={Guo, Dong and Wu, Faming and Zhu, Feida and Leng, Fuxing and Shi, Guang and Chen, Haobin and Fan, Haoqi and Wang, Jian and Jiang, Jianyu and Wang, Jiawei and others},
  journal={arXiv preprint arXiv:2505.07062},
  year={2025}
}

@article{team2025kimi,
  title={Kimi-vl technical report},
  author={Team, Kimi and Du, Angang and Yin, Bohong and Xing, Bowei and Qu, Bowen and Wang, Bowen and Chen, Cheng and Zhang, Chenlin and Du, Chenzhuang and Wei, Chu and others},
  journal={arXiv preprint arXiv:2504.07491},
  year={2025}
}

@article{seedream2025seedream,
  title={Seedream 4.0: Toward next-generation multimodal image generation},
  author={Seedream, Team and Chen, Yunpeng and Gao, Yu and Gong, Lixue and Guo, Meng and Guo, Qiushan and Guo, Zhiyao and Hou, Xiaoxia and Huang, Weilin and Huang, Yixuan and others},
  journal={arXiv preprint arXiv:2509.20427},
  year={2025}
}

@article{team2025openai,
    title ={Thinking with images},
    author ={OpenAI},
    year ={2025},
    journal={https://openai.com/index/thinking-with-images/}
}

@article{team2025okuaishou,
    title ={Kolors2.0},
    author ={Kuaishou Kolors team},
    year ={2025},
    journal={https://app.klingai.com/cn/}
}

@article{chen2025janus,
  title={Janus-pro: Unified multimodal understanding and generation with data and model scaling},
  author={Chen, Xiaokang and Wu, Zhiyu and Liu, Xingchao and Pan, Zizheng and Liu, Wen and Xie, Zhenda and Yu, Xingkai and Ruan, Chong},
  journal={arXiv preprint arXiv:2501.17811},
  year={2025}
}

@article{xie2025reconstruction,
  title={Reconstruction alignment improves unified multimodal models, 2025a},
  author={Xie, Ji and Darrell, Trevor and Zettlemoyer, Luke and Wang, XuDong},
  journal={URL https://arxiv. org/abs/2509.07295},
  volume={1},
  pages={3},
  year={2025}
}

@article{liao2025mogao,
  title={Mogao: An omni foundation model for interleaved multi-modal generation},
  author={Liao, Chao and Liu, Liyang and Wang, Xun and Luo, Zhengxiong and Zhang, Xinyu and Zhao, Wenliang and Wu, Jie and Li, Liang and Tian, Zhi and Huang, Weilin},
  journal={arXiv preprint arXiv:2505.05472},
  year={2025}
}

@article{jiang2025t2i,
  title={T2i-r1: Reinforcing image generation with collaborative semantic-level and token-level cot},
  author={Jiang, Dongzhi and Guo, Ziyu and Zhang, Renrui and Zong, Zhuofan and Li, Hao and Zhuo, Le and Yan, Shilin and Heng, Pheng-Ann and Li, Hongsheng},
  journal={arXiv preprint arXiv:2505.00703},
  year={2025}
}

@article{deng2025emerging,
  title={Emerging properties in unified multimodal pretraining},
  author={Deng, Chaorui and Zhu, Deyao and Li, Kunchang and Gou, Chenhui and Li, Feng and Wang, Zeyu and Zhong, Shu and Yu, Weihao and Nie, Xiaonan and Song, Ziang and others},
  journal={arXiv preprint arXiv:2505.14683},
  year={2025}
}

@article{yu2022scaling,
  title={Scaling autoregressive models for content-rich text-to-image generation},
  author={Yu, Jiahui and Xu, Yuanzhong and Koh, Jing Yu and Luong, Thang and Baid, Gunjan and Wang, Zirui and Vasudevan, Vijay and Ku, Alexander and Yang, Yinfei and Ayan, Burcu Karagol and others},
  journal={arXiv preprint arXiv:2206.10789},
  volume={2},
  number={3},
  pages={5},
  year={2022}
}

@inproceedings{hu2023tifa,
  title={Tifa: Accurate and interpretable text-to-image faithfulness evaluation with question answering},
  author={Hu, Yushi and Liu, Benlin and Kasai, Jungo and Wang, Yizhong and Ostendorf, Mari and Krishna, Ranjay and Smith, Noah A},
  booktitle={Proceedings of the IEEE/CVF International Conference on Computer Vision},
  pages={20406--20417},
  year={2023}
}

@article{huang2023t2i,
  title={T2i-compbench: A comprehensive benchmark for open-world compositional text-to-image generation},
  author={Huang, Kaiyi and Sun, Kaiyue and Xie, Enze and Li, Zhenguo and Liu, Xihui},
  journal={Advances in Neural Information Processing Systems},
  volume={36},
  pages={78723--78747},
  year={2023}
}

@article{ghosh2023geneval,
  title={Geneval: An object-focused framework for evaluating text-to-image alignment},
  author={Ghosh, Dhruba and Hajishirzi, Hannaneh and Schmidt, Ludwig},
  journal={Advances in Neural Information Processing Systems},
  volume={36},
  pages={52132--52152},
  year={2023}
}

@article{hu2024ella,
  title={Ella: Equip diffusion models with llm for enhanced semantic alignment},
  author={Hu, Xiwei and Wang, Rui and Fang, Yixiao and Fu, Bin and Cheng, Pei and Yu, Gang},
  journal={arXiv preprint arXiv:2403.05135},
  year={2024}
}

@article{li2024genai,
  title={Genai-bench: Evaluating and improving compositional text-to-visual generation},
  author={Li, Baiqi and Lin, Zhiqiu and Pathak, Deepak and Li, Jiayao and Fei, Yixin and Wu, Kewen and Ling, Tiffany and Xia, Xide and Zhang, Pengchuan and Neubig, Graham and others},
  journal={arXiv preprint arXiv:2406.13743},
  year={2024}
}

@article{wu2024conceptmix,
  title={Conceptmix: A compositional image generation benchmark with controllable difficulty},
  author={Wu, Xindi and Yu, Dingli and Huang, Yangsibo and Russakovsky, Olga and Arora, Sanjeev},
  journal={Advances in Neural Information Processing Systems},
  volume={37},
  pages={86004--86047},
  year={2024}
}

@article{huang2025t2i,
  title={T2i-compbench++: An enhanced and comprehensive benchmark for compositional text-to-image generation},
  author={Huang, Kaiyi and Duan, Chengqi and Sun, Kaiyue and Xie, Enze and Li, Zhenguo and Liu, Xihui},
  journal={IEEE Transactions on Pattern Analysis and Machine Intelligence},
  volume={47},
  number={5},
  pages={3563--3579},
  year={2025},
  publisher={IEEE}
}

@misc{zhuo2025factualitymattersimagegeneration,
      title={Factuality Matters: When Image Generation and Editing Meet Structured Visuals}, 
      author={Le Zhuo and Songhao Han and Yuandong Pu and Boxiang Qiu and Sayak Paul and Yue Liao and Yihao Liu and Jie Shao and Xi Chen and Si Liu and Hongsheng Li},
      year={2025},
      eprint={2510.05091},
      archivePrefix={arXiv},
      primaryClass={cs.CV},
      url={https://arxiv.org/abs/2510.05091}, 
}

@misc{chang2025oneigbenchomnidimensionalnuancedevaluation,
      title={OneIG-Bench: Omni-dimensional Nuanced Evaluation for Image Generation}, 
      author={Jingjing Chang and Yixiao Fang and Peng Xing and Shuhan Wu and Wei Cheng and Rui Wang and Xianfang Zeng and Gang Yu and Hai-Bao Chen},
      year={2025},
      eprint={2506.07977},
      archivePrefix={arXiv},
      primaryClass={cs.CV},
      url={https://arxiv.org/abs/2506.07977}, 
}

@misc{geng2025xomnireinforcementlearningmakes,
      title={X-Omni: Reinforcement Learning Makes Discrete Autoregressive Image Generative Models Great Again}, 
      author={Zigang Geng and Yibing Wang and Yeyao Ma and Chen Li and Yongming Rao and Shuyang Gu and Zhao Zhong and Qinglin Lu and Han Hu and Xiaosong Zhang and Linus and Di Wang and Jie Jiang},
      year={2025},
      eprint={2507.22058},
      archivePrefix={arXiv},
      primaryClass={cs.CV},
      url={https://arxiv.org/abs/2507.22058}, 
}

@article{li2026gebench,
  title={GEBench: Benchmarking Image Generation Models as GUI Environments},
  author={Haodong Li and Jingwei Wu and Quan Sun and Guopeng Li and Juanxi Tian and Huanyu Zhang and Yanlin Lai and Ruichuan An and Hongbo Peng and Yuhong Dai and Chenxi Li and Chunmei Qing and Jia Wang and Ziyang Meng and Zheng Ge and Xiangyu Zhang and Daxin Jiang},
  journal={arXiv preprint arXiv:2602.09007},
  year={2026}
}

@misc{zhang2026vibe-benchmark,
  title={How Well Do Models Follow Visual Instructions? VIBE: A Systematic Benchmark for Visual Instruction-Driven Image Editing}, 
      author={Huanyu Zhang and Xuehai Bai and Chengzu Li and Chen Liang and Haochen Tian and Haodong Li and Ruichuan An and Yifan Zhang and Anna Korhonen and Zhang Zhang and Liang Wang and Tieniu Tan},
      year={2026},
      eprint={2602.01851},
      archivePrefix={arXiv},
      primaryClass={cs.CV},
      url={https://arxiv.org/abs/2602.01851}, 
}

@misc{an2026geniusgenerativefluidintelligence,
      title={GENIUS: Generative Fluid Intelligence Evaluation Suite}, 
      author={Ruichuan An and Sihan Yang and Ziyu Guo and Wei Dai and Zijun Shen and Haodong Li and Renrui Zhang and Xinyu Wei and Guopeng Li and Wenshan Wu and Wentao Zhang},
      year={2026},
      eprint={2602.11144},
      archivePrefix={arXiv},
      primaryClass={cs.LG},
      url={https://arxiv.org/abs/2602.11144}, 
}

@article{wang2026deepgen,
  title={DeepGen 1.0: A Lightweight Unified Multimodal Model for Advancing Image Generation and Editing},
  author={Wang, Dianyi and Li, Ruihang and Han, Feng and Ma, Chaofan and Song, Wei and Wang, Siyuan and Wang, Yibin and Xin, Yi and Liu, Hongjian and Zhang, Zhixiong and others},
  journal={arXiv preprint arXiv:2602.12205},
  year={2026}
}

@article{han2026unicorn,
      title={UniCorn: Towards Self-Improving Unified Multimodal Models through Self-Generated Supervision}, 
      author={Han, Ruiyan and Fang, Zhen and Sun, XinYu and Ma, Yuchen and Wang, Ziheng and Zeng, Yu and Chen, Zehui and Chen, Lin and Huang, Wenxuan and Xu, Wei-Jie and others},
      journal={arXiv preprint arXiv:2601.03193},
      year={2026},
}

@article{wei2025tiif,
  title={TIIF-Bench: How Does Your T2I Model Follow Your Instructions?},
  author={Wei, Xinyu and Zhang, Jinrui and Wang, Zeqing and Wei, Hongyang and Guo, Zhen and Zhang, Lei},
  journal={arXiv preprint arXiv:2506.02161},
  year={2025}
}

@inproceedings{li20251+,
  title={Why $1+ 1 < 1$ in Visual Token Pruning: Beyond Naive Integration via Multi-Objective Balanced Covering},
  author={Li, Yangfu and Zhan, Hongjian and Chen, Tianyi and Liu, Qi and Xiong, Yu-Jie and Lu, Yue},
  booktitle={The Thirty-ninth Annual Conference on Neural Information Processing Systems},
  year={2025}
}

@inproceedings{lyu2024unibind,
  title={Unibind: Llm-augmented unified and balanced representation space to bind them all},
  author={Lyu, Yuanhuiyi and Zheng, Xu and Zhou, Jiazhou and Wang, Lin},
  booktitle={Proceedings of the IEEE/CVF Conference on Computer Vision and Pattern Recognition},
  pages={26752--26762},
  year={2024}
}

@article{lyu2025realrag,
  title={Realrag: Retrieval-augmented realistic image generation via self-reflective contrastive learning},
  author={Lyu, Yuanhuiyi and Zheng, Xu and Jiang, Lutao and Yan, Yibo and Zou, Xin and Zhou, Huiyu and Zhang, Linfeng and Hu, Xuming},
  journal={arXiv preprint arXiv:2502.00848},
  year={2025}
}

@article{lyu2025understanding,
  title={Understanding-in-Generation: Reinforcing Generative Capability of Unified Model via Infusing Understanding into Generation},
  author={Lyu, Yuanhuiyi and Wong, Chi Kit and Liao, Chenfei and Jiang, Lutao and Zheng, Xu and Lu, Zexin and Zhang, Linfeng and Hu, Xuming},
  journal={arXiv preprint arXiv:2509.18639},
  year={2025}
}

@inproceedings{lideepscan,
      title={DeepScan: A Training-Free Framework for Visually Grounded Reasoning in Large Vision-Language Models}, 
      author={Yangfu Li and Hongjian Zhan and Jiawei Chen and Yuning Gong and Qi Liu and Yue Lu},
      booktitle={IEEE/CVF conference on computer vision and pattern recognition},
      year={2026},
}

@article{lin2025perceive,
  title={Perceive anything: Recognize, explain, caption, and segment anything in images and videos},
  author={Lin, Weifeng and Wei, Xinyu and An, Ruichuan and Ren, Tianhe and Chen, Tingwei and Zhang, Renrui and Guo, Ziyu and Zhang, Wentao and Zhang, Lei and Li, Hongsheng},
  journal={arXiv preprint arXiv:2506.05302},
  year={2025}
}

@article{li2026videococo,
  title={VideoCoCo: Code-as-CoT for Physically-Consistent Video Generation via an Agentic Dual-Engine System},
  author={Li, Haodong and Ren, Tianfei and Ma, Xiaoxiao and Qing, Chunmei and Fang, Zhen and He, Sipeng and Guo, Ziyu and Wu, Haoyu and Tian, Juanxi and Zou, Yihang and others},
  journal={arXiv preprint arXiv:2607.27380},
  year={2026}
}

@article{tian2026auto,
  title={Auto-rubric as reward: From implicit preferences to explicit multimodal generative criteria},
  author={Tian, Juanxi and Liu, Fengyuan and Han, Jiaming and Jiang, Yilei and Wu, Yongliang and Liu, Yesheng and Li, Haodong and Xu, Furong and Li, Wanhua},
  journal={arXiv preprint arXiv:2605.08354},
  year={2026}
}

@article{yan2026proact,
  title={Proact-VL: A Proactive VideoLLM for Real-Time AI Companions},
  author={Yan, Weicai and Dai, Yuhong and Ran, Qi and Li, Haodong and Lin, Wang and Jin, Tao and Xie, Xing and Liao, Hao and Lian, Jianxun},
  journal={arXiv preprint arXiv:2603.03447},
  year={2026}
}

@article{wei2026perceptionrubrics,
  title={PerceptionRubrics: Calibrating Multimodal Evaluation to Human Perception},
  author={Wei, Yana and Peng, Hongbo and Lai, Yanlin and Zhao, Liang and Lin, Kangheng and Yu, En and Lv, Keyu and Zhou, Han and Tang, Yin and Li, Haodong and others},
  journal={arXiv preprint arXiv:2606.28322},
  year={2026}
}

\end{document}